%% file: main_camera_ready.tex
\documentclass[journal]{IEEEtai}

\usepackage[colorlinks,urlcolor=blue,linkcolor=blue,citecolor=blue]{hyperref}

\usepackage{color,array}

\usepackage{graphicx}

\usepackage{amsmath}
\usepackage{amssymb}
\usepackage{booktabs}
\usepackage{algorithm}
\usepackage{algorithmic}
\usepackage{mathrsfs}
\usepackage{xcolor}
\usepackage{multirow}
\usepackage{colortbl}
\usepackage{epsfig}

\newcommand{\eg}{\textit{e}.\textit{g}.}
\newcommand{\etal}{\textit{et al}.}
\newcommand{\ie}{\textit{i}.\textit{e}.}

\definecolor{light_green}{HTML}{E3F2D9}
\definecolor{deep_green}{HTML}{C8E5B3}
\definecolor{deep_pink}{HTML}{FB31A2}
\usepackage{hyperref}
\hypersetup{urlcolor=deep_pink,
            }

\begin{document}

\title{Spiking Diffusion Models} 

\author{Jiahang Cao$^{*}$\thanks{$^*$ Equal contribution. $\ddagger$ Second author. $\dagger$ Corresponding author.}, Hanzhong Guo$^{*}$, Ziqing Wang$^{*}$, Deming Zhou$^{\ddagger}$, Hao Cheng, Qiang Zhang, and Renjing Xu$^{\dagger}$
\thanks{Jiahang Cao, Deming Zhou, Hao Cheng, and Qiang Zhang are with the Hong Kong University of Science and Technology (Guangzhou), Guangzhou, China. Email: \href{mailto:jcao248@connect.hkust-gz.edu.cn}{jcao248@connect.hkust-gz.edu.cn}.}
\thanks{Hanzhong Guo is with the Renmin University of China, Beijing, China. Email: \href{mailto:guohanzhong@ruc.edu.cn}{guohanzhong@ruc.edu.cn}.}
\thanks{Ziqing Wang is with the North Carolina State University, North Carolina, USA. Email: \href{mailto:zwang247@ncsu.edu}{zwang247@ncsu.edu}.}
\thanks{Renjing Xu is with the Hong Kong University of Science and Technology (Guangzhou), Guangzhou, China. Email: \href{mailto:renjingxu@hkust-gz.edu.cn}{renjingxu@hkust-gz.edu.cn}.}
\thanks{© 20XX IEEE.  Personal use of this material is permitted.  Permission from IEEE must be obtained for all other uses, in any current or future media, including reprinting/republishing this material for advertising or promotional purposes, creating new collective works, for resale or redistribution to servers or lists, or reuse of any copyrighted component of this work in other works.}}


\maketitle

\begin{abstract}

Recent years have witnessed Spiking Neural Networks (SNNs) gaining attention for their ultra-low energy consumption and high biological plausibility compared with traditional Artificial Neural Networks (ANNs). Despite their distinguished properties, the application of SNNs in the computationally intensive field of image generation is still under exploration. In this paper, we propose the Spiking Diffusion Models (SDMs), an innovative family of SNN-based generative models that excel in producing high-quality samples with significantly reduced energy consumption. 
In particular, we propose a Temporal-wise Spiking Mechanism (TSM) that allows SNNs to capture more temporal features from a bio-plasticity perspective. In addition, we propose a threshold-guided strategy that can further improve the performances by up to 16.7\% without any additional training.
We also make the first attempt to use the ANN-SNN approach for SNN-based generation tasks.
Extensive experimental results reveal that our approach not only exhibits comparable performance to its ANN counterpart with few spiking time steps, but also outperforms previous SNN-based generative models by a large margin. Moreover, we also demonstrate the high-quality generation ability of SDM on large-scale datasets, \eg, LSUN bedroom. 
This development marks a pivotal advancement in the capabilities of SNN-based generation, paving the way for future research avenues to realize low-energy and low-latency generative applications. Our code is available at \url{https://github.com/AndyCao1125/SDM}.
\end{abstract}

\begin{IEEEImpStatement}
Diffusion models have achieved great success in image synthesis through iterative noise estimation using deep neural networks. However, the slow inference, high memory consumption, and computation intensity of the noise estimation model hinder the efficient adoption of diffusion models. In this paper, we overcame these challenges by introducing spiking diffusion models. 
In particular, the SDM showcased substantial energy efficiency, consuming merely $\sim$30\% of the energy required by the ANN model, while still delivering superior generative outcomes. This technology could offer an alternative way of achieving sustainable, low-energy, and efficient image generation tasks.
\end{IEEEImpStatement}


\vspace{10mm}
\section{Introduction}

\begin{figure}[t]
\begin{center}
\vspace{3pt}
\begin{tabular}{@{}c@{}}
\includegraphics[width=0.85\linewidth]{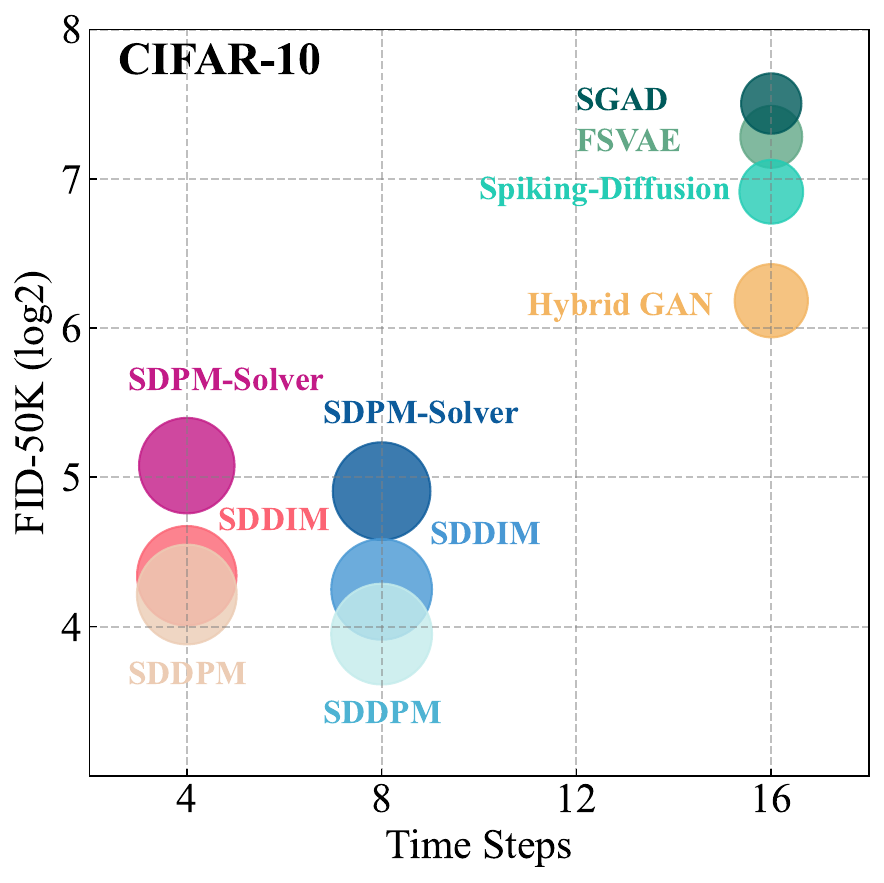} \\
\end{tabular}

\end{center}
\caption{\textbf{Comparisons of the state-of-the-art SNN models.} The FID is at $log_2$ scale and the marker size corresponds to the IS metric. In comparison to other SNN generative models, our models demonstrate better FID while requiring lower time steps.}
\label{fig:compare_start}
\end{figure}
\IEEEPARstart{H}{uman} brain is recognized as an amazingly intelligent system that regulates the connections between multiple neuronal layers to cope with a variety of complex tasks. Inspired by this, Artificial Neural Networks (ANNs) mimic the structural patterns of the human brain, achieving performance that approaches human levels across various fields (\eg, GPT4~\cite{achiam2023gpt}). However, the great success of artificial neural networks is highly dependent on a large number of intensive computations~\cite{sze2017efficient}, whereas the human brain only requires a small power budget ($\approx$20 Watts)~\cite{raichle2002appraising,cox2014neural} and performs operations in an asynchronous and low-latency manner. 

In the quest for the next wave of advanced intelligent systems, Spiking Neural Networks (SNNs), being regarded as the third generation of neural networks, are gaining interest for their distinctive attributes: high biological plausibility, event-driven nature, and low power consumption. 
In SNNs, all information is represented in the form of spikes (binary vectors that can either be 0 or 1) diverging from the continuous representation in ANNs, which allows SNNs to adopt spike-based low-power accumulation (AC) instead of the traditional high-power multiply-accumulation (MAC), thus achieving considerable improvements in energy efficiency. Existing works reveal that employing SNNs on neuromorphic hardware, such as Loihi~\cite{davies2018loihi}, TrueNorth~\cite{akopyan2015truenorth} and Tianjic~\cite{pei2019towards} can save energy by orders of magnitude compared with ANNs.
Due to the specialties of SNNs, they have been widely adopted in various tasks, 
including recognition~\cite{zhou2022spikformer,deng2022temporal}, tracking~\cite{zhang2022spiking}, detection~\cite{kim2020spiking,yuan2024trainable}, segmentation~\cite{kirkland2020spikeseg} and images restoration~\cite{li2021event}.

Recently, deep generative models (DGMs) and especially (score-based) diffusion models~\cite{sohl2015deep,ho2020denoising,song2020score,karras2022elucidating,cao2024spiking} have made remarkable progress in various domains, including text-to-image generation~\cite{ho2022cascaded, dhariwal2021diffusion,xu2022versatile,bao2023one}, audio generation~\cite{kong2020diffwave,popov2021grad}, video generation~\cite{ho2022video}, text-to-3D  generation~\cite{poole2022dreamfusion,wang2024prolificdreamer}. 
However, the main challenge is that the computational cost of the diffusion model is high, due to the computational complexity of the ANN backbone and the large number of iterations of the denoising process.

Hence, leveraging SNN is a potential way to develop energy-efficient diffusion models. However, SNN-based diffusion models remain little studied. Liu \etal~\cite{liu2023spiking} introduce Spiking-Diffusion based on the vector quantized variational autoencoder (VQ-VAE), where image features are encoded using spike firing rate and diffusion process is performed in the discrete latent space to generate images. Kamata \etal~\cite{kamata2022fully} propose a fully spiking variational autoencoder (FSVAE) combined with discrete Bernoulli sampling.
Despite these advancements, the image produced by these SNN-based diffusion models tends to lag behind that of existing generation models~\cite{ho2020denoising,nichol2021improved}. 
Consequently, it is imperative to develop a generative algorithm capable of producing high-quality images while also reducing energy consumption.

In this work, we propose Spiking Diffusion Models (SDMs), a new family of SNN-based diffusion models with excellent image generation capability and low energy consumption. SDMs can be applied to any diffusion solvers, including but not limited to DDPM~\cite{ho2020denoising}, DDIM~\cite{song2020denoising}, Analytic-DPM~\cite{bao2022analytic} and DPM-solver~\cite{lu2022dpm}. Our approach adopts the pre-spike spiking UNet structure, ensuring the spikes can be transmitted more accurately. In addition, we propose Temporal-wise Spiking Mechanism (TSM), a bio-plasticity mechanism that enables the membrane potential to be dynamically updated at each time step to capture more time-dependent features. We also provide training-free threshold guidances, which further enhance the quality of resulting images by adjusting the threshold value of the spiking neurons. 
Experimental results demonstrate that both inhibitory and excitatory guidance provide facilitation for SDMs. 
Moreover, we make the \textit{first attempt} to use the ANN-SNN method for providing SDMs with comparable results.
We evaluate our approach on both standard datasets (\eg, CIFAR-10, and CelebA) and large-scale datasets (\eg, LSUN bedroom). As depicted in Fig.~\ref{fig:compare_start}, our proposed SDMs outperform all SNN-based generative models by a substantial margin with only a few spiking time steps.

In summary, our contributions are:
\begin{itemize}
    \item We propose Spiking Diffusion Model, a high-quality image generator that achieves state-of-the-art performances among SNN-based generative models and can be applied to any diffusion solvers.
    \item From a biological perspective, we propose a Temporal-wise Spiking Mechanism (TSM), allowing spiking neurons to capture more dynamic information to enhance the quality of denoised images.
    \item Extensive results show SDMs outperform the SNN baselines by up to 12$\times$ of the FID score on the CIFAR-10 while saving $\sim$60\% energy consumption.  
    Also, we put forward a threshold-guidance strategy to further improve the generative performance.
\end{itemize}

\begin{figure*}[t!]
	\setlength{\tabcolsep}{1.0pt}
	\centering
	\begin{tabular}{c}
        \includegraphics[width=0.8\textwidth]{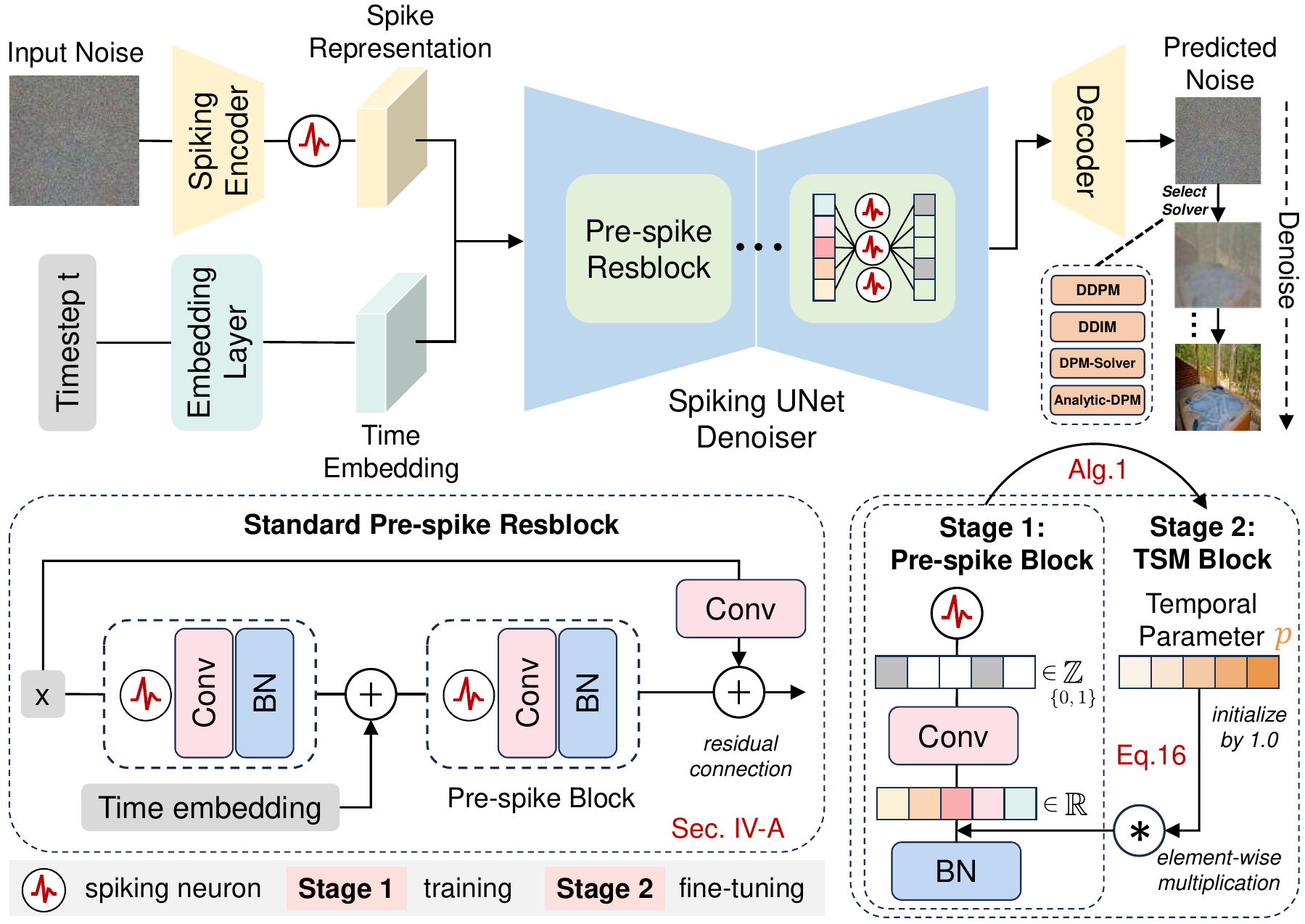} 
	\end{tabular}
	\caption{\textbf{Overview of our Spiking Diffusion Models.} The learning process of SDM consists of two stages: (1) the training stage and (2) the fine-tuning stage. During the training stage, our spiking UNet adopts the standard Pre-spike Resblock (bottom left, Sec.~\ref{subsec:pre_spike}), and then converts the Pre-spike block into the TSM block (bottom right, Sec.~\ref{subsec:tsm}) for the fine-tuning stage. Given a random Gaussian noise input $x_t$, it is firstly converted into the spike representation by a spiking encoder and subsequently fed into the spiking UNet along with the time embeddings. The network transmits only spikes which are represented by $0/1$ vector ($\in \mathbb{Z}_{\{0,1\}}$). 
 Finally, the output spikes are passed through a decoder to obtain the predicted noise $\epsilon$, 
 and the loss is computed to update the network. In the fine-tuning phase, we load the weights from the training phase and substitute the Pre-spike block with the TSM block, where the temporal parameter $p$ is initialized as 1.0. This stage continues to optimize the network's parameters for better generative performance.
 }
	\label{fig:pipeline}
\end{figure*}

\section{Related Work}

\subsection{Training Methods of Spiking Neural Networks}
Generally, ANN-to-SNN conversion and direct training are two mainstream ways to obtain deep SNN models:

\noindent\textbf{Direct Training.}
Direct training methods involve training the SNN directly from scratch. The inherent challenge with direct training stems from the non-differentiability of the spiking neuron's output, which complicates the application of traditional backpropagation techniques reliant on continuous gradients. To address this challenge, direct training methods utilize surrogate gradients~\cite{neftci2019surrogate, lee2020enabling, xiao2021training} for backpropagation. These surrogate gradients offer a smooth approximation of the spiking function, facilitating the use of gradient-based optimization methods for SNN training.

\noindent\textbf{ANN-to-SNN.} 
The basic idea of the ANN-to-SNN conversion methods~\cite{bu2023optimal,deng2021optimal,ding2021optimal,li2021free} 
is to replace the ReLU activation values in ANNs with the average firing rates of spiking neurons. The ANN-to-SNN conversion method is generally known to achieve higher accuracy compared to direct training methods. However, they typically require a longer training time compared to direct training methods, as the conversion process requires more time steps and additional optimization which may restrict the practical SNN application.

In our study, we delve into the application of direct training methods to integrate diffusion models within SNN frameworks, aiming to reduce power consumption and investigate the potential generative abilities of SNNs. In addition, we extend our investigation to include the ANN-to-SNN conversion method for implementing SNN-based diffusion models, allowing for a comprehensive comparison of outcomes between the two training paradigms.

\subsection{Spiking Neural Network in Generative Models}
Most of the SNN-based generation algorithms mainly originate from the generation model in ANNs, such as VAEs and GANs.~\cite{skatchkovsky2021learning, stewart2022encoding, rosenfeld2022spiking} propose hybrid architectures consisting an SNN-based encoder and an ANN-based decoder. However, these approaches utilize the structure of the ANN, leading to the entire model could not be fully deployed on neuromorphic hardware. Spiking GAN~\cite{kotariya2022spiking} adopts a fully SNN-based backbone and employs a time-to-first-spike coding scheme. It significantly increases the sparsity of the spike series, therefore giving large energy savings. Kamata \etal~\cite{kamata2022fully} then propose a fully spiking variational autoencoder (FSVAE) that is able to transmit only spikes throughout the whole generation process sampled by the Bernoulli distribution. Feng \etal~\cite{feng2023sgad} construct a spiking GAN with attention scoring decoding (SGAD) and identify the problems of out-of-domain inconsistency and temporal inconsistency inherent, which increases the performance compared to existing spiking GAN-based methods. Recently, Watanabe \etal~\cite{watanabe2023fully} propose a fully spiking denoising diffusion implicit model to achieve high speed and low energy consumption features of SNNs via synaptic current learning.

However, irrespective of whether the proposed models are based on VAE/GAN, or entirely based on spiking neurons, the primary limitation of existing spiking generative models is their low performance and poor quality of generated images. These drawbacks hinder their competitiveness in the field of generative models, despite their low energy consumption. To tackle this issue, we introduce SDMs, which not only deliver substantial improvements over existing SNN-based generative models but also preserve the advantages of SNNs.

\section{Background}
\subsection{Spiking Neural Network}
The Spiking Neural Network is a bio-inspired algorithm that mimics the actual signaling process occurring in brains.
Compared to the Artificial Neural Network (ANN), it transmits sparse spikes instead of continuous representations, offering benefits such as low energy consumption and robustness. 
In this paper, we adopt the widely used Leaky Integrate-and-Fire (LIF~\cite{hunsberger2015spiking,burkitt2006review}) model, which effectively characterizes the dynamic process of spike generation and can be defined as:
\begin{align}
    & U[n] = e^{\frac{1}{\tau}}V[n-1] + I[n] \label{eq:dis_lif1},\\
    & S[n] = \Theta (U[n] - \vartheta_{\textrm{th}})\label{eq:dis_lif2},\\
    & V[n] = U[n](1-S[n]) + V_{\textrm{reset}}S[n] \label{eq:dis_lif3},
\end{align}
where $n$ is the time step, $U[n]$ is the membrane potential before reset, $S[n]$ denotes the output spike which equals 1 when there is a spike and 0 otherwise, $\Theta(x)$ is the Heaviside step function, $V[n]$ represents the membrane potential after triggering a spike. In addition, we use the “hard reset” method~\cite{fang2021incorporating} for resetting the membrane potential in Eq.~\eqref{eq:dis_lif3}, which means that the value of the membrane potential $V[n]$ after triggering a spike ($S[n]=1$) will go back to $V_{\textrm{reset}} = 0$. 

\subsection{Diffusion Models and Classifier-Free Guidance}
\label{subsec:back_diff}
Diffusion models leverage a forward and reverse process for data generation. They gradually perturb data with a forward diffusion process, then learn to reverse this process to recover the data distribution. The process is detailed as follows:

Formally, let $x_0\in\mathbb{R}^n$ be a random variable with unknown data distribution $q(x_0)$. 
The forward diffusion process $\left \{ x_t \right \} _{t\in [0,T]}$ indexed by time $t$, can be represented by the following forward Stochastic differential equations~(SDE):
\begin{eqnarray}
\label{eq:sdef}
     \textrm{d} x_t = f(t)x_t\textrm{d} t + g(t)\textrm{d} \omega  ,\quad x_0\sim q(x_0),
\end{eqnarray}
where $\omega \in \mathbb{R}^n$ is a standard Wiener process. 
Let $q(x_t)$ be the marginal distribution of the above SDE at time $t$. Its reversal process can be described by a corresponding continuous SDE which recovers the data distribution~\cite{song2020score}:
\begin{eqnarray}
\label{eq:continuous_back}
    \textrm{d} x = \left [ f(t)x_t-\frac{1+\lambda^2}{2}g^2(t) \nabla_{x_t} \log q(x_t) \right ] \textrm{d} t + \lambda g(t)\textrm{d} \bar{\omega} ,
\end{eqnarray}
where $\bar{\omega} \in \mathbb{R}^n$ is a reverse-time standard Wiener process, $\lambda$ controls the andomness added while keeping the marginal distribution $q(x_t)$ the same, and this general reversed SDE starts from $x_T\sim q(x_T)$. 
In Eq.~\eqref{eq:continuous_back}, the only unknown term is the score function $\nabla_{x_t} \log q(x_t)$. To estimate this term, prior works~\cite{ho2020denoising, song2020score, karras2022elucidating} employ a noise network $\epsilon_{\theta}(x_t,t)$ to obtain a scaled score function $\sigma(t)\nabla_{x_t} \log q(x_t)$ using denoising score matching~\cite{vincent2011connection}, which ensures that the optimal solution equals to the ground truth and satisfies $\epsilon_{\theta}(x_t,t)=-\sigma(t) \nabla_{x_t} \log q(x_t)$, where $\alpha(t),\sigma(t)$ represents the noise schedule and $x_t$ are sample from $q(x_t|x_0) \sim \mathcal{N}(x_t|a(t)x_0,\sigma^2(t)I)$.
\begin{eqnarray}
\label{eq:gab}
    f(t)=\frac{\textrm{d} \log a(t)}{\textrm{d} t} ,\quad g^2(t)=\frac{\textrm{d} \sigma^2(t)}{\textrm{d} t}-2\sigma^2(t)\frac{\textrm{d} \log a(t)}{\textrm{d} t} .
\end{eqnarray}

Hence, sampling can be achieved by discretizing the reverse SDE 
Eq.~\eqref{eq:continuous_back} when replacing the $\nabla_{x_t} \log q(x_t)$ with noise network.
What's more, in order to achieve conditional sampling, we can refine the reverse SDE in Eq.~\eqref{eq:continuous_back} and let $\lambda=1$ as:
\begin{eqnarray}
\label{eq:condition}
    &\textrm{d} x = \left [ f(t)x_t-g^2(t)\frac{(1+\omega )\epsilon_{\theta}(x_t)-\omega\epsilon_{\theta}(x_t,c)}{\sigma(t)} \right ] \textrm{d} t + g(t)\textrm{d} \bar{\omega}  \nonumber\\
& \epsilon _{\theta}(x_t,c) = \epsilon _{\theta}(x_t)-\sigma(t)\nabla_{x_t}\log p(c|x_t),
\end{eqnarray}
where $\log p(c|x_t)$ is the output probability of classifier and the second equation is derived from the Bayesian formula which implies that a conditional sample can be obtained without extra training. 
The term $(1+\omega)\epsilon_{\theta}(x_t)-\omega\epsilon_{\theta}(x_t,c)$ represents classifier-free guidance, which effectively enhances the diversity of generated samples. This approach has been widely adopted in text-to-image diffusion models~\cite{IF}, as evidenced by the works of Ho et al.~\cite{ho2022classifier}.

Furthermore, it is important to note that the guidance mentioned above is not limited to a specific classifier. It can be applied to various forms of guidance function. For example, energy-based guidance~\cite{zhao2022egsde, bao2022equivariant} is proposed to achieve image translation and molecular design under a guidance of energy-based function. 
Additionally, Kim et al.~\cite{kim2022refining} introduce discriminator guidance to mitigate estimation bias in the noise network, resulting in state-of-the-art performance on the CIFAR-10 dataset.

\section{Method}

\subsection{Pre-spike Residual Learning}
\label{subsec:pre_spike}

We first analyze the limitations and conceptual inconsistencies present in the residual learning approaches of previous Spiking Neural Networks (SNNs), specifically SEW ResNet~\cite{fang2021deep}, which can be formulated as:
\begin{align}
    O^l &= \mathrm{BN}^l(\mathrm{Conv}^l(S^{l-1})) + S^{l-1},\\
    S^l &= \mathrm{SpikeNeuron}(O^{l}),\\
    O^{l+1} &= \mathrm{BN}^{l+1}(\mathrm{Conv}^{l+1}(S^{l})) + S^{l},\\
    S^{l+1} &= \mathrm{SpikeNeuron}(O^{l+1}),
\end{align}
where $O^l$ represents the output after the batch normalization (BN) and convolutional operations at the $l^{th}$ layer, and $S^l$ denotes the spiking neuron activation function in Eq.~\eqref{eq:dis_lif2}. Such a residual structure~\cite{zheng2021going,hu2021spiking,fang2021deep} is inherited directly from traditional artificial neural network (ANN) ResNet architectures~\cite{he2016deep}. However, a fundamental issue arises with this approach concerning the output range of the residual blocks. The core of the problem lies in the nature of the spiking neuron outputs ($S^{l-1}$ and $S^l$), which are binary spike sequences, taking values in the set $\{0,1\}$. Consequently, when these sequences are summed in the residual structure ($O^l$), the resulting summation output domain expands to $\{0,1,2\}$. The occurrence of the value $\{2\}$ in this context is non-biological and represents a departure from plausible neural activation patterns. This overflow condition not only detracts from the biological realism of the model but also potentially disrupts the effective transmission of information during forward propagation.

Inspired by~\cite{liu2018bi,zhang2022pokebnn}
, we adopt pre-spike residual learning with the structure of $Activation$-$Conv$-$BatchNorm$
in our Spiking UNet, addressing the dual challenges of gradient explosion/vanishing and performance degradation in convolution-based SNNs. Through the pre-spike
blocks, the residuals and outputs are summed by floating point addition operation, ensuring that the representation is accurate before entering the next spiking neuron while avoiding the pathological condition mentioned above. The whole pre-spike residual learning process inside a resblock can be formulated as below:
\begin{align}
    S^{l} &= \mathrm{SpikeNeuron}(O^{l-1}),\\
    O^{l} &= \mathrm{BN}^l(\mathrm{Conv}^l(S^l)) + O^{l-1},\\
    S^{l+1} &= \mathrm{SpikeNeuron}(O^{l}),\\
    O^{l+1} &= \mathrm{BN}^{l+1}(\mathrm{Conv}^{l+1}(S^{l+1})) + O^{l}.
\end{align}
Through the pre-spike residual mechanism, the output of the residual block can be summed by two floating points $BN^l(Conv^l(S^l))$, $O^{l-1}$ at the same scale and then enter the spiking neuron at the beginning of the next block, which guarantees that the energy consumption is still very low.

\subsection{Temporal-wise Spiking Mechanism }
\label{subsec:tsm}

In this section, we first revisit the shortcomings of traditional SNNs from a biological standpoint and propose a novel TSM mechanism that fine-tunes the weights to capture temporal dynamics by incorporating temporal parameters.
Considering the spike input at $l$-th layer as $S^l \in \mathbb{R}^{N\times T\times C_{in}\times H\times W}$, where $N$ denotes the minibatch size. For every time step $n\in \{1,2,...,T\}$, the neuron will update its temporary membrane potential via Eq.~\eqref{eq:dis_lif1},
where $I^{l-1}[n] = \mathbf{W}^lS^{l-1}[n]$ and $\mathbf{W}^l \in \mathbb{R}^{N\times C_{out}\times H\times W}$ denotes the weight matrix of the $l$-th convolutional layer. 
The traditional SNNs~\cite{zheng2021going} will fuse the $T$ and $C$ dimension of the input into $\widetilde{S}^l\in \mathbb{R}^{NT\times C_{in}\times H\times W}$ when performing the membrane potential update, and then compute it by 2D convolution operation. This leads to the input of each time step being operated by the same weight matrix. However, in the real nervous system, cortical pyramidal cells receive strong barrages of both excitatory and inhibitory postsynaptic potentials during regular network activity~\cite{hasenstaub2005inhibitory}. Moreover, different states of arousal can alter the membrane potential and impact synaptic integration~\cite{steriade2001natural}. These studies collectively demonstrate that the neuron input at each moment experiences considerable fluctuations due to the network states and other factors, rather than being predominantly controlled by fixed synaptic weights.

\begin{figure}[t]
\begin{center}
\vspace{3pt}
\begin{tabular}{@{}c@{}}
\includegraphics[width=\linewidth]{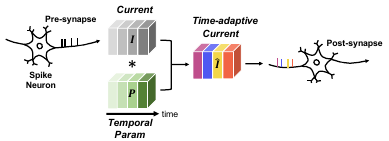} \vspace{-3mm} \\
\end{tabular}
\vspace{-2mm}
\end{center}
\caption{\textbf{Overview of temporal-wise spiking mechanism.} After a spike neuron triggers spikes, the spikes would be converted in the pre-synapse to obtain the input current $I$. To derive more dynamic information, the temporal parameter $P$ will act on the current to form the time-adaptive current $\hat{I}$.}
\label{figure:tsm}
\vspace{-3pt}
\end{figure}

In order to provide more dynamic information through time, we propose the Temporally-wise Spiking Mechanism (TSM, see Fig.~\ref{figure:tsm}), which guarantees that the input information at each moment is computed with a temporal parameter $p[n]\in \mathbf{P}$ ($\mathbf{P}\in \mathbb{R}^{T}$) related to time step $n$:
\begin{align}
    U^l[n] = e^{\frac{1}{\tau}}V^l[n-1] + (\mathbf{W}^l S^{l-1}[n])\times p[n]
\end{align}

We describe the whole learning process of SDMs 
precisely in Alg.~\ref{alg:pipeline}. 
Specifically, the learning pipeline consists of (1) \textit{the training stage} and (2) \textit{the fine-tuning stage}, where we first train the SDMs with pre-spike block and fine-tune the model by utilizing TSM block. We only use few iterations $I_{ft}$ ($< \frac{1}{10} $ training iterations) to fine-tune our model. It is worth noting that, since $p[t]$ is only a scalar, \textit{the extra computational costs incurred by TSM are negligible}. By calculating $\mathcal{L}_{MSE}$, we could further optimize the parameters of $\mathbf{W}$ and $\mathbf{P}$ to obtain satisfying networks. The detailed learning rules of SNN-UNet and $p[t]$ can be found in Appendix.

In summary, TSM allows the membrane potential to be dynamically updated across the time domain, improving the ability to capture potential time-dependent features. Subsequent experiments demonstrate that the TSM mechanism is superior to the traditional fixed update mechanism. We also visualized $\mathbf{P}$ for detailed analysis in Sec.~\ref{subsec:tsm_method}.

\vspace{-10pt}
\subsection{Threshold Guidance in SDMs}
\label{subsec:tg}

As mentioned in Sec.~\ref{subsec:back_diff}, the sampling can be achieved by substituting the score $\nabla_{x_t} \log q(x_t)$ with either the score network $s_{\theta}(x_t,t)$ or the scaled noise network -$\frac{\epsilon_{\theta}(x_t,t)}{\sigma(t)}$ while discretizing the reverse SDE as presented in Eq.~\eqref{eq:continuous_back}. 
Because of the inaccuracy of the network estimates, we have the fact that $s_{\theta}(x_t,t) \approx -\frac{\epsilon_{\theta}(x_t,t)}{\sigma(t)} \ne \nabla_{x_t} \log q(x_t)$ in most cases. Therefore, in order to sample better results, we can discretize the following rectified reverse SDE: 
\begin{eqnarray}
\label{eq:rec_continuous_back}
    \textrm{d} x = \left [ f(t)x_t-g^2(t) [s_{\theta}+c_{\theta}](x_t,t)\right ] \textrm{d} t + g(t)\textrm{d} \bar{\omega} ,
\end{eqnarray}
Here, $s_{\theta}(x_t,t)$ denotes the score network or scaled noise network, while $c_{\theta}(x_t,t)=\nabla_{x_t} \log \frac{q(x_t)}{p_{\theta}(x_t,t)}$ represents the rectified term for the original reverse SDE. The omission of the rectified term $c_{\theta}(x_t,t)$ results in reduced discretization error and improved sampling performance. However, the practical calculation of $c_{\theta}(x_t,t)$ poses challenges due to its intractability.

Given the presence of estimation error, Eq.~\eqref{eq:rec_continuous_back} prompts us to investigate whether we can enhance sampling performance without additional training by calculating $c_{\theta}(x_t,t)$. A crucial parameter in the SNN is the spike threshold, denoted as $\vartheta_{\textrm{th}}$, which directly influences the SNN's output. For instance, a smaller threshold encourages more spike occurrences, while a larger threshold suppresses such occurrences.
Following the training process, we can adjust the threshold within the SNN to estimate the rectified term $c_{\theta}(x_t,t)$ as outlined below:
\begin{align}
\label{eq:taylor}
&s_{\theta}(x_t,t,\vartheta_{\textrm{th}}') \nonumber \\
&\approx s_{\theta}(x_t,t,\vartheta_{\textrm{th}}^0)+\frac{\mathrm{d} s_{\theta}(x_t,t,\vartheta_{\textrm{th}})}{\mathrm{d} \vartheta_{\textrm{th}}} \mathrm{d} \vartheta_{\textrm{th}} + O(\mathrm{d} \vartheta_{\textrm{th}}) \nonumber \\
&\approx s_{\theta}(x_t,t)+ s'_{\theta}|_{\vartheta_{\textrm{th}}^0} \mathrm{d} \vartheta_{\textrm{th}} + O(\mathrm{d} \vartheta_{\textrm{th}})
 \nonumber \\
&\approx s_{\theta}(x_t,t)+c_{\theta}(x_t,t) ,
\end{align}
where $\vartheta_{\textrm{th}}^0$ is the threshold used in the training stage and $\vartheta_{\textrm{th}}'$ is the adjusted threshold used in the inference stage. The first equation is obtained by Talyor expansion. Eq.~\eqref{eq:taylor} tells that when the derivative term is related to the rectified term, adjusting the threshold can boost the final sampling results. We call the threshold-decreasing case as the inhibitory guidance, and conversely as excitatory guidance. 

\input{Alg_tsm.tex}

\section{Theoretical Energy Consumption Calculation}
\label{subsec:energy}

In this section, we describe the methodology for calculating the theoretical energy consumption of the Spiking UNet architecture. The calculation involves two main steps: determining the synaptic operations (SOPs) for each block within the architecture, and then estimating the overall energy consumption based on these operations. The synaptic operations for each block of the Spiking UNet can be quantified as follows~\cite{zhou2022spikformer}:
\begin{equation}
  \operatorname{SOPs}(l)=fr \times T \times \operatorname{FLOPs}(l)  
\end{equation}
where $l$ denotes the block number in the Spiking UNet, $fr$ is the firing rate of the input spike train of the block and $T$ is the time step of the spike neuron. $\operatorname{FLOPs}(l)$ refers to floating point operations of $l$ block, which is the number of multiply-and-accumulate (MAC) operations. And SOPs are the number of spike-based accumulate $(\mathrm{AC})$ operations. 

To estimate the theoretical energy consumption of Spiking Diffusion, we assume that the MAC and AC operations are implemented on a $45 nm$ hardware, with energy costs of $E_{MAC} = 4.6 pJ$ and $E_{AC} = 0.9 pJ$, respectively. According to~\cite{panda2020toward, yao2023attention}, the calculation for the theoretical energy consumption of Spiking Diffusion is given by:
\begin{equation}
 \begin{aligned}
E_{\text {Diffusion}} & =E_{MAC} \times \mathrm{FLOP}_{\mathrm{SNN}_\mathrm{Conv}}^1 \\
& +E_{AC} \times\left(\sum_{n=2}^N \mathrm{SOP}_{\mathrm{SNN}_\mathrm{Conv}}^n+\sum_{m=1}^M \mathrm{SOP}_{\mathrm{SNN}_\mathrm{FC}}^m\right)
\end{aligned}   
\end{equation}
where $N$ and $M$ represent the total number of layers of convolutional (Conv) and fully connected (FC) layers, respectively. $E_{MAC}$ and $E_{AC}$ energy costs per operation for MAC and AC, respectively. $\mathrm{FLOP}_{\mathrm{SNN}_\mathrm{Conv}}$ refers to the FLOPs of the first Conv layer, and $\mathrm{SOP}_{\mathrm{SNN}_\mathrm{Conv}}$ and $\mathrm{SOP}_{\mathrm{SNN}_\mathrm{FC}}$ are the SOPs for the $n^{th}$ Conv and $m^{th}$ FC layers, respectively.

\section{Implement Spiking Diffusion Models via ANN-SNN conversion}
\label{subsec:ann-snn}

In this paper, \textit{for the first time}, we also utilize the ANN-SNN approach to successfully implement SNN diffusion. we adopt the Fast-SNN~\cite{hu2023fast} approach to construct the conversion between quantized ANNs and SNNs. Since this implementation is not the main contribution of our paper, we briefly describe the ANN-SNN principle and more details can be found in~\cite{hu2023fast}.

The core idea of the conversion from ANNs to SNNs is to map the integer activation of quantized ANNs $\left\{0, 1, \dots, 2^b-1\right\}$ to spike count $\left\{0, 1, \dots, T\right\}$, \ie, convert $T$ to $2^b-1$. Building quantized ANNs with integer activations is naturally equivalent to compressing activations with the quantization function that outputs uniformly distributed values. Such a function spatially discretizes a full-precision activation $x^l_i$ of neuron $i$ at layer $l$ in an ANN of ReLU activation into:     
\begin{equation}\label{eq:2}
	Q^l_i=\dfrac{s^l}{2^b-1}\mathrm{clip}(\mathrm{round}((2^b-1) \dfrac{x^l_i}{s^l}),0, 2^b-1),
\end{equation}
where $Q^l_i$ denotes the spatially quantized value, $b$ denotes the number of bits (precision), the number of states is $2^b-1$, $\mathrm{round}(\cdot)$ denotes a rounding operator, $s^l$ denotes the clipping threshold that determines the clipping range of input $x^l_i$, $\mathrm{clip}(x, min, max)$ is a clipping operator that saturates $x$ within range $[min, max]$. 

In SNNs, the spiking IF neuron inherently quantizes the membrane potential $U^l_i$ into a quantized value represented by the firing rate $r^l_i$:

\begin{equation}\label{eq:7}
	\widetilde{Q}^l_i=r^l_i=\dfrac{1}{T}\mathrm{clip}(\mathrm{floor}(\dfrac{U^l_i}{\theta^l}),0,T),
\end{equation}
where $\widetilde{Q}^l_i$ denotes the spiking-based quantized value, $\mathrm{floor}(\cdot)$ denotes a flooring operator. Assume that the value of membrane potential is always satisfied to be $T$ times the value of the input current: $U^l_i = T I^l_i$. Comparing Eq.~\ref{eq:7} with Eq.~\ref{eq:2}, we let $\mu^l = \theta^l/2$, $T = 2^b-1$, $\theta^l=s^l$. 
Since a flooring operator can be converted to a rounding operator:
\begin{equation}\label{eq:9}
	\mathrm{floor}(x+0.5) = \mathrm{round}(x),
\end{equation}
By scaling the weights in the following layer to $s^l W^{l+1}$, we rewrite Eq.~\ref{eq:7} into the following equation :
\begin{equation}\label{eq:11}
	\widetilde{Q}^l_i=\dfrac{s^l}{T}\mathrm{clip}(\mathrm{floor}(\dfrac{TI^l_i}{\theta^l}),0,T)= Q^l_i.
\end{equation}
Hence, by establishing the equivalence of discrete ReLU value and spike firing rate (Eq.~\ref{eq:11}), we build a bridge between quantized ANNs and SNNs. It is important to note that the assumption of $U^l_i = T I^l_i$ only holds in the first spiking layer which directly receives currents as inputs. However, as the network goes deeper, the interplay between membrane potential and input current grows increasingly complex, deviating from a simple linear relationship. This complexity is one of the fundamental reasons for the progressively larger error accumulation in the ANN-SNN conversion process.

\section{Experiment}

\subsection{Experiment Settings}
\noindent\textbf{Datasets and Evaluation Metrics.}
To demonstrate the effectiveness and efficiency of the proposed algorithm, we conduct experiments on 32$\times$32 MNIST~\cite{lecun2010mnist}, 32$\times$32 Fashion-MNIST~\cite{xiao2017fashion}, 32$\times$32 CIFAR-10~\cite{krizhevsky2009learning} and 64$\times$64 CelebA~\cite{liu2015deep}. The qualitative results are compared according to Fr\'{e}chet Inception Distance (FID~\cite{heusel2017gans}, lower is better) and Inception Score (IS~\cite{salimans2016improved}, higher is better). FID score is computed by comparing 50,000
generated images against the corresponding reference statistics of the dataset.

\noindent\textbf{Implementation Details.} 
For the direct training method, our Spiking UNet inherits the standard UNet~\cite{ronneberger2015u} architecture and 
no attention blocks are used. 
For the hyper-parameter settings, we set the decay rate $e^{\frac{1}{\tau}}$ in Eq.~\eqref{eq:dis_lif1} as 1.0 and the spiking threshold $\vartheta_{
\textrm{th}}$ as 1.0. The SNN simulation time step is 4/8. The learning rate is set as 1
e-5 with batch size 128 and we train the model
without exponential moving average (EMA~\cite{tarvainen2017mean}). ANN UNet also does not employ attention blocks, and its training process is consistent with SNN-UNet. For the ANN-SNN method, we use the same implementation of Fast-SNN~\cite{hu2023fast}, but we do not apply the signed-IF neuron since this neuron plays a negative role in the diffusion task. More details of the hyperparameter settings can be found in the Appendix.

\begin{table*}[t]\scriptsize
\caption{\textbf{Results for different dataset.}
In all datasets, SDMs (Ours) outperform all SNN baselines 
and even some ANN models in terms of sample quality, which is mainly measured by FID and IS. Results of $^\triangledown$ are taken from \cite{kamata2022fully} and results of $^\natural$ are taken from \cite{feng2023sgad}. $_\textit{ema}$ indicates the utilization of EMA~\cite{tarvainen2017mean} method. For fair comparisons, we re-evaluate the results of DDPM~\cite{ho2020denoising} and DDIM~\cite{song2020denoising} using the same UNet architecture as SDMs. $^*$ denotes that only FID is used for MNIST, Fashion-MNIST and CelebA since their data distributions are far from ImageNet, making Inception Score less meaningful. The top-1 and top-2 results are bold and underlined, respectively.}
\label{tab:mainresults}
\setlength\tabcolsep{12pt} 
\centering
\resizebox{.95\linewidth}{!}{
\begin{tabular}{ccccccc}
\toprule 
\multirow{1}{*}{\textbf{Dataset}} &
  \multirow{1}{*}{\textbf{Model}} &
  \multirow{1}{*}{\textbf{Method}} &
  \multirow{1}{*}{\textbf{\#Param (M)}} &
  \multirow{1}{*}{\textbf{Time Steps}} &
  \multirow{1}{*}{\textbf{IS}$\uparrow$} &
  \multirow{1}{*}{\textbf{FID}$\downarrow$} \\ \midrule
\multirow{5}{*}{MNIST$^*$} 
& VAE$^\triangledown$~\cite{kingma2013auto} & ANN &1.13 & {/} & 5.947 & 112.50 \\
& Hybrid GAN$^\natural$~\cite{rosenfeld2022spiking} & SNN\&ANN &-&16 &- &123.93\\
\cmidrule{2-7}
& FSVAE~\cite{kamata2022fully} &SNN &3.87 & 16 & 6.209 & 97.06        \\
& SGAD~\cite{feng2023sgad} &SNN & -& 16 & - & 69.64 \\
& \multicolumn{1}{c}{{Spiking-Diffusion~\cite{liu2023spiking}}} &{SNN} & {-}& {16} & {-} & {37.50} \\           
& \textbf{SDDPM} &SNN& 63.61 & 4 & - & \textbf{29.48}\\\midrule
\multirow{5}{*}{\begin{tabular}[c]{@{}c@{}}Fashion\\ MNIST$^*$\end{tabular}} 
& VAE~\cite{kingma2013auto} & ANN &1.13 & {/} & 4.252 & 123.70 \\
& Hybrid GAN~\cite{rosenfeld2022spiking} & SNN\&ANN &-&16 &- &198.94\\
\cmidrule{2-7}
& FSVAE~\cite{kamata2022fully} &SNN & 3.87 & 16 &  4.551 & 90.12  \\
& SGAD~\cite{feng2023sgad} &SNN &-& 16 & - & 165.42 \\ 
& \multicolumn{1}{c}{{Spiking-Diffusion~\cite{liu2023spiking}}} &{SNN} & {-}& {16} & {-} & {91.98} \\
& \textbf{SDDPM} &SNN&63.61& 4 & - & \textbf{21.38} 
\\ \midrule
\multirow{2}{*}{{LSUN bedroom$^*$}} 
& {DDPM~\cite{ho2020denoising}} & {ANN} & {64.47} & {/} & {-}& {29.48}\\
\cmidrule{2-7}      
 & {\textbf{SDDPM}} &{SNN} & {63.61} & {4} &  {-} & {47.64}\\ \midrule
\multirow{7}{*}{CelebA$^*$} 
& VAE~\cite{kingma2013auto} & ANN &3.76 & {/} & 3.231 & 92.53 \\
& Hybrid GAN~\cite{rosenfeld2022spiking} & SNN\&ANN &-&16 &- &63.18\\ 
& DDPM~\cite{ho2020denoising} & ANN & 64.47 & {/} & -& 20.34\\
\cmidrule{2-7}
 & FSVAE~\cite{kamata2022fully} &SNN& 6.37 & 16 & 3.697 & 101.60          \\ 
 & SGAD~\cite{feng2023sgad} &SNN&-& 16 & - & 151.36 \\  
 & FSDDIM~\cite{watanabe2023fully} & SNN & - & 4 & - & 36.08 \\
 & \textbf{SDDPM} &SNN & 63.61 & 4 &  - & \textbf{25.09}\\ \midrule
 \multirow{17}{*}{CIFAR-10}  
 & VAE~\cite{kingma2013auto} & ANN &1.13 & {/} & 2.591 & 229.60 \\
& Hybrid GAN~\cite{rosenfeld2022spiking} & SNN\&ANN &-&16 &- &72.64\\
& DDIM~\cite{song2020denoising} & ANN & 64.47 &{/} &8.428 &18.49 \\
& DDPM~\cite{ho2020denoising} & ANN & 64.47 &{/} &8.380 &19.04 \\
& DDIM$_\textit{ema}$~\cite{song2020denoising} & ANN & 64.47 &{/} &8.902 &12.13 \\
& DDPM$_\textit{ema}$~\cite{ho2020denoising} & ANN & 64.47 &{/} &8.846 &13.38 \\
\cmidrule{2-7}
& FSVAE~\cite{kamata2022fully} &SNN & 3.87 & 16 & 2.945 & 175.50   \\
& SGAD~\cite{feng2023sgad} &SNN& - & 16 & - & 181.50 \\
& \multicolumn{1}{c}{{Spiking-Diffusion~\cite{liu2023spiking}}} &{SNN} & {-}& {16} & {-} & {120.50} \\
& FSDDIM~\cite{watanabe2023fully} & SNN & - & 4 & - & 51.46 \\
& FSDDIM~\cite{watanabe2023fully} & SNN & - & 8 & - & 46.14 \\
&\textbf{SDDPM} &SNN&  63.61& 4 & 7.440 & 19.73\\
&\textbf{SDDPM} &SNN& 63.61 & 8 & 7.584 & 17.27\\
&\textbf{SDDPM (TSM)} &SNN& 63.61 (+2$e^{-4}$) & 4 & 7.654 & 18.57\\
&\textbf{SDDPM (TSM)} &SNN& 63.61 (+2$e^{-4}$)& 8 & 7.814 & \underline{15.45}\\
&\textbf{SDDIM (TSM)} &SNN& 63.61 (+2$e^{-4}$)& 8 & \underline{7.834} & 16.02\\
&\textbf{SDPM-Solver (TSM)} &SNN& 63.61 (+2$e^{-4}$) & 8 & 7.180 & 30.85\\
&\textbf{Analytic-SDPM (TSM)} &SNN& 63.61 (+2$e^{-4}$) & 8 &\textbf{7.844}  &\textbf{12.95} \\
\bottomrule
\end{tabular}}
\end{table*}

\subsection{Comparisons with the state-of-the-art}

In Tab.~\ref{tab:mainresults}, we present a comparative analysis of our Spiking Diffusion Models (SDMs) with state-of-the-art generative models in unconditional generations.  We also include ANN results for reference.  Qualitative results are shown in Fig.~\ref{fig:visualization}.
Our results demonstrate that \textit{SDMs outperforms SNN baselines across all datasets by a significant margin}, even with smaller spiking simulation steps (4/8). 
In particular, SDDPM has 4$\times$ and 6$\times$ FID improvement in CelebA, while 11$\times$ and 12$\times$ enhancement in CIFAR-10 compared to FSVAE and SGAD (both are with 16 time steps). As expected, the sample quality becomes higher as we increase the time step. We additionally note that the incorporation of TSM leads to enhance performance with only a negligible increase in model parameters (2$e^{-4}$ M).
SDMs can also handle fast sampling solvers~\cite{bao2022analytic,lu2022dpm} and attain higher sampling quality in fewer steps (see Tab.~\ref{tab:fast_solver}). Importantly, SDMs obtain comparable quality to ANN baselines with the same UNet architecture and \textit{even surpasses ANN models}(e.g., 15.45 vs.~19.04). 
This outcome highlights the superior expressive capability of SNNs employed in our model.

\begin{table}[t]
\setlength\tabcolsep{10pt} 
\centering
\caption{\textbf{Comparisons of diffusion models with direct training and ANN-SNN conversion.} While ANN-SNN approaches always outperform direct training methods in classification tasks, their effectiveness falls short when applied to generative tasks, where direct training yields superior results.}
	\label{tab:ann-snn}
\resizebox{1\linewidth}{!}{
\begin{tabular}{lcccc}
\toprule
\textbf{Model}&\textbf{Method} & \textbf{Time Steps}       & \textbf{IS}$\uparrow$  & \textbf{FID}$\downarrow$     \\ \midrule
SDDPM         & Direct Training    &4   & 7.645          & 18.57 \\
SDDPM         & Direct Training    &8   & \textbf{7.814}          & \textbf{15.45} \\
SDDPM (w/o ft)  & ANN-SNN   & 3 & 5.484 & 51.18 \\
SDDPM (w ft)  & ANN-SNN   & 3 & 7.223 & 29.53
\\ \bottomrule       
\end{tabular}}
\end{table}

\subsection{Comparisons with the ANN-SNN method}
To validate the generative ability of SDM under the ANN-SNN approach, we conduct experiments on the 32$\times$32 CIFAR-10 and 64$\times$64 FFHQ~\cite{karras2019style} datasets. As shown in Tab.~\ref{tab:ann-snn}, the ANN-SNN approach achieves great performance on CIFAR-10 (\ie, 51.18 FID) and improves the image quality substantially after the fine-tuning strategy (\ie, 29.53 FID). However, there is still a gap between the results of ANN-SNN and those of direct training. Although ANN-SNN methods are demonstrated to have comparable performances to ANN's in classification-based tasks, there is still a lack of in-depth research on generative tasks. The qualitative results of the ANN-SNN method are illustrated in Fig.~\ref{fig:ann2snn}.

\begin{figure}[t]
\centering
\setlength{\tabcolsep}{0pt} 
\renewcommand{\arraystretch}{0.8} 
\begin{tabular}{c}
  \includegraphics[width=0.45\textwidth]{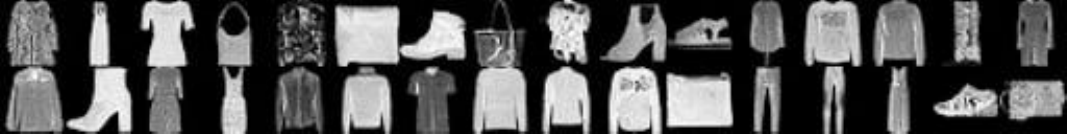} \\
  \includegraphics[width=0.45\textwidth]{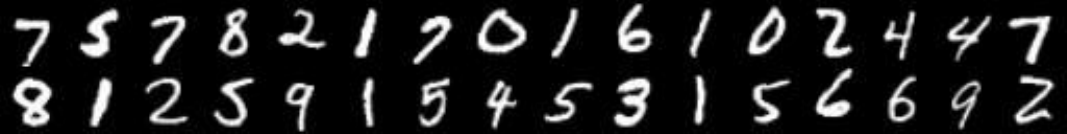} \\
  \includegraphics[width=0.45\textwidth]{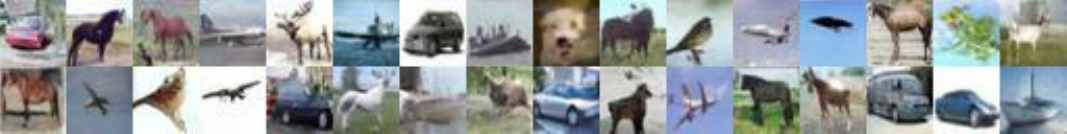} \\
  \includegraphics[width=0.45\textwidth]{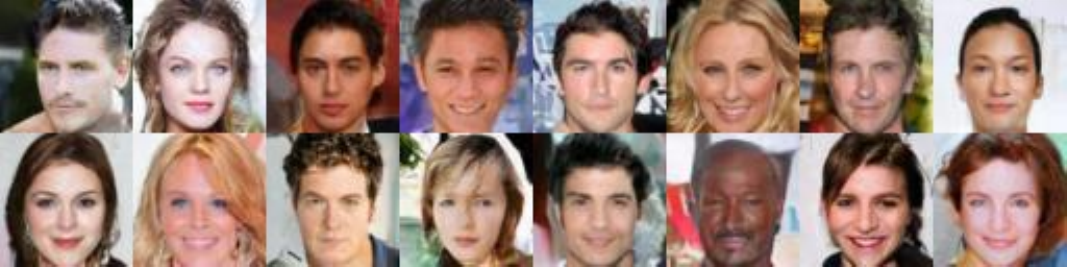} \\
  \includegraphics[width=0.45\textwidth]{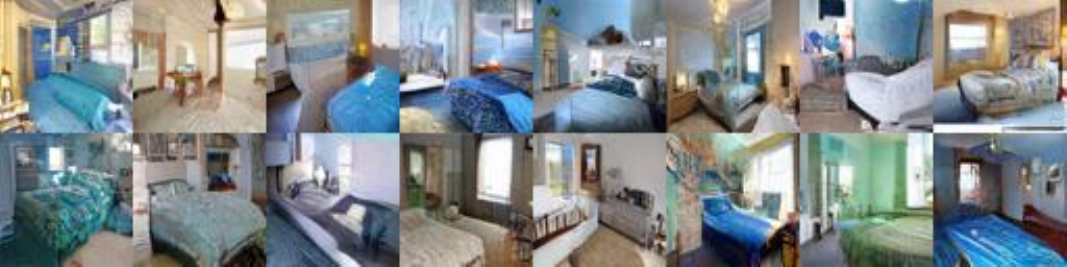}
\end{tabular}
\caption{\textbf{Unconditional image generation results on MNIST, Fashion-MNIST, CIFAR-10, CelebA, and LSUN-bed by using direct training-based SDMs.}}
\label{fig:visualization}
\end{figure}

\begin{table}[t]
        \setlength\tabcolsep{20pt}
        \centering
        \caption{\textbf{Results on CIFAR-10 by different threshold guidances.} 
        The top-1 and top-2 results are bold and underlined, respectively.}
        \label{tab:TG}
        \resizebox{1\linewidth}{!}{
        \begin{tabular}{cccc}
        \toprule 
    	 \textbf{Method} &\textbf{Threshold} & \textbf{FID}$\downarrow$ & \textbf{IS}$\uparrow$   \\
        \midrule
        Baseline& 1.000   & 19.73 &  7.44  \\
                            \midrule
        \multirow{3}{*}{\shortstack{Inhibitory\\ Guidance}}         & 0.999  & {\underline{19.25}}  & {\underline{7.48}}   \\
                            & 0.998  & 19.38   & \textbf{{7.55}}  \\
                             & 0.997  & \textbf{{19.20}}  &  7.47  \\
                             \midrule
        \multirow{3}{*}{\shortstack{Excitatory\\ Guidance}} & 1.001  & 20.00  &  7.47 \\
                             & 1.002  & 19.98  &  {\underline{7.48}} \\
                             & 1.003  & 20.04   & 7.46  \\ 
                            \bottomrule
        \end{tabular}}
\end{table}

\subsection{Effectiveness of the Temporal-wise Spiking Mechanism}
To better visualize the performance improvement brought by the TSM module, we provide the generation results of CIFAR-10 using SDDIM with and without the TSM module. Here we use DDIM~\cite{song2020denoising} instead of DDPM~\cite{ho2020denoising} for this comparison since DDIM operates based on Ordinary Differential Equations (ODEs), which ensure deterministic and consistent generation results. In contrast, DDPM relies on Stochastic Differential Equations (SDEs), which introduce randomness in the generation process, leading to variability in the output images and making direct comparisons challenging.

\begin{figure}[!t]
	\setlength{\tabcolsep}{1.0pt}
	\centering
	\begin{tabular}{c}
        \includegraphics[width=.49\textwidth]{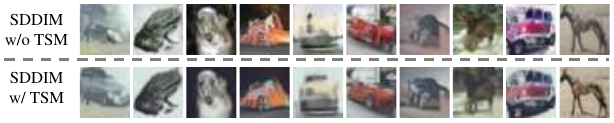} 
	\end{tabular}
        \vspace{-4mm}
	\caption{\textbf{Comparisons of the generation results with/without using the TSM method in CIFAR-10.}
 }
	\label{fig:vis}
\end{figure}

The results in Fig.~\ref{fig:vis} demonstrate a significant improvement in the quality of the generated images with the TSM module. The contours of the images are more pronounced, the backgrounds are clearer, and the texture details are richer compared to those without the TSM module, thereby proving the effectiveness of TSM. 

\subsection{Effectiveness of Threshold Guidance}
In Sec.~\ref{subsec:tg}, we propose a training-free method: Threshold Guidance (TG), designed to enhance the quality of generated images by merely adjusting the threshold levels of spiking neurons slightly during the inference phase. As depicted in Tab.~\ref{tab:TG},  
applying inhibitory guidance through threshold adjustment significantly elevates image quality across two key metrics: the FID score decreases from 19.73 to 19.20 with a threshold reduction of 0.3\%, and the IS score climbs from 7.44 to 7.55 following a 0.2\% threshold decrease.
Conversely, excitatory guidance similarly augments the sampling quality under certain conditions. 
These findings underscore the potency of threshold guidance as a means to substantially improve model efficacy post-training, without necessitating extra training resources. We provide more explanations about the threshold guidance in the Appendix.

\begin{table}[t]
\setlength\tabcolsep{10pt} 
\centering
\caption{\textbf{Comparisons of energy and FID of SNN and ANN models.} In comparison to ANN, SNN models demonstrate lower energy consumption while achieving comparable FID results.}
\label{tab:energy}
\resizebox{1\linewidth}{!}{
\begin{tabular}{cccc}
\toprule
\textbf{Model}      & DDPM-ANN & SDDPM-4T       & SDDPM-8T       \\ \midrule
\textbf{FID$\downarrow$}         & 19.04    & 18.57          & \textbf{15.45} \\
\textbf{Energy (mJ)$\downarrow$} & 29.23    & \textbf{10.97} & 22.96  \\ \bottomrule       
\end{tabular}}
        
\end{table}

\subsection{Analysis of TSM method}
\label{subsec:tsm_method}
\begin{figure}[t]
\begin{center}
\begin{tabular}{@{}c@{}}
\includegraphics[width=0.8\linewidth]{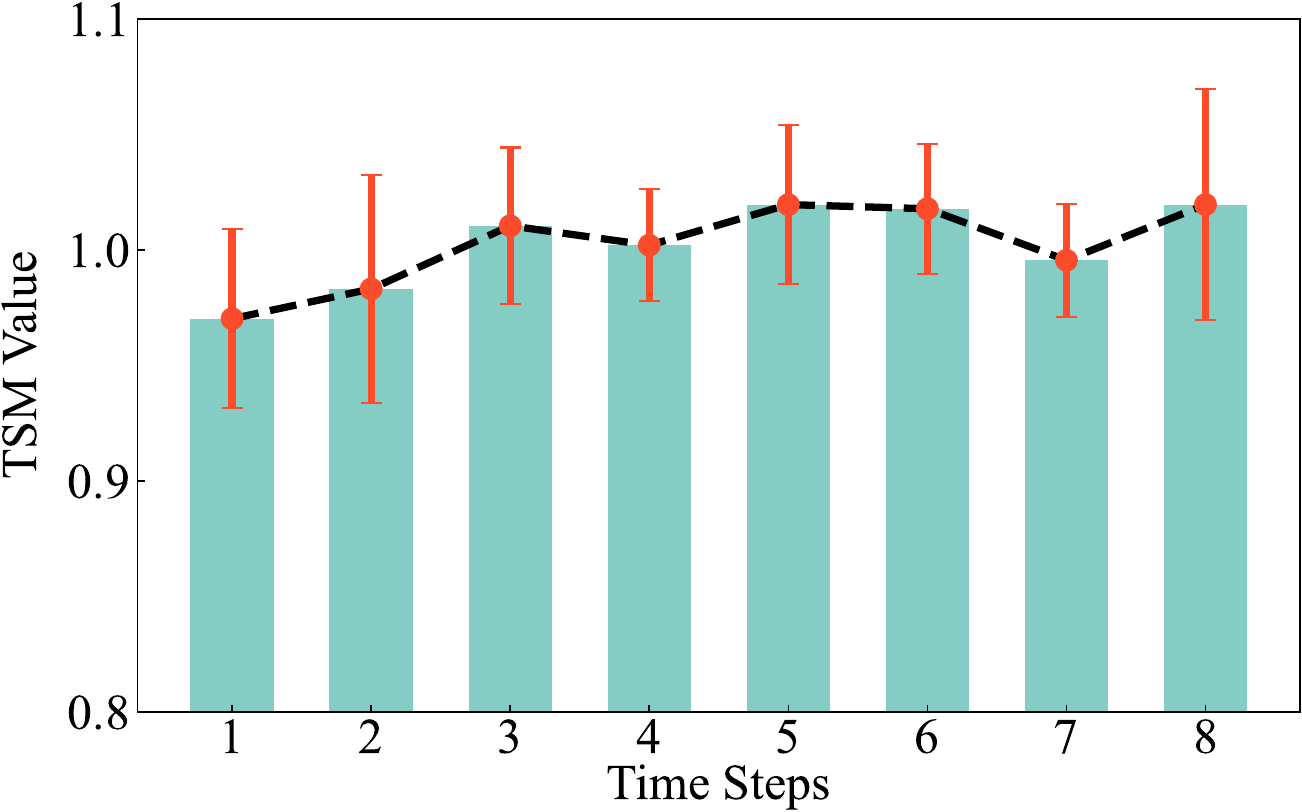} \vspace{-3mm} \\
\end{tabular}
\end{center}
\caption{\textbf{Visualization of TSM Value.} The averaged TSM values across all layers are depicted, with the red error bar.}
\label{fig:tsm_vis}
\end{figure}

To evaluate the efficacy of the Temporal-wise Spiking Mechanism (TSM) proposed in Sec.~\ref{subsec:tsm}, we compute the mean of the temporal parameters over all layers. As depicted in Fig.~\ref{fig:tsm_vis}, each instance presents distinct TSM values $ p[t]$, which underscores the unique significance attributed to each time step. We notice that the evolutionary trend of $p[t]$ exhibits an increasing pattern as the number of time steps grows, suggesting that the information conveyed in later stages holds greater importance during the transmission process.
As a result, the TSM values can serve as temporal adjusting factors, enabling the SNN to comprehend and incorporate temporal dynamics, thus improving the quality of the generated image.

\begin{table}[b]
    \caption{\textbf{Ablation study on different proposed methods.} The experiments are conducted on SDDIM (T=4) with 50 denoising steps. $\Delta$  represents the improvement of FID. }
    \label{tab:ablation}
    \renewcommand{\arraystretch}{1.1}
        \setlength\tabcolsep{14pt} 
        \centering
        \resizebox{1\linewidth}{!}{
        \begin{tabular}{ccc|cc}
        \toprule 
    	 \textbf{Solver} &\textbf{TSM} & \textbf{TG} & \textbf{FID}$\downarrow$ & $\Delta$ (\%)   \\
        \hline
        \multirow{4}{*}{\shortstack{SDDIM}}
        &  &   &20.68 (-0.00)  & +0.00  \\
        & \checkmark  &   &20.26 (-0.42)  &+2.03  \\
        &   & \checkmark   & 19.62 (-1.06)   &+5.12\\
        &  \checkmark & \checkmark  & \textbf{16.88 (-3.80)} &+18.4   \\ 
                            \bottomrule
        \end{tabular}}
        
\end{table}

\subsection{Evaluation of the Computational Cost}

To further emphasize the low-energy nature of our SDMs, we perform a comparative analysis of the FID and energy consumption between the proposed SDDPM and its corresponding ANN model. As shown in Tab.~\ref{tab:energy}, when the time step is set at 4, the SDDPM presents significantly lower energy consumption, amounting to merely 37.5\% of that exhibited by its ANN counterpart. Moreover, the FID of SDDPM also improved by 0.47, indicating that our model can effectively minimize energy consumption while maintaining competitive performance. As we extend the analysis to include varying time step increments, a discernible pattern emerges: the FID score improves as the time step increases, albeit at the expense of higher energy consumption. This observation points to a trade-off between FID improvement and the associated energy expenses as time steps increase.

\begin{figure}[t]
\centering
\setlength{\tabcolsep}{0pt}
\renewcommand{\arraystretch}{0.8}
\begin{tabular}{c}
  \includegraphics[width=0.45\textwidth]{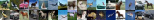} \\
  \includegraphics[width=0.45\textwidth]{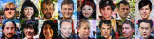}
\end{tabular}
\caption{\textbf{Unconditional image generation results on CIFAR-10 and FFHQ64 by using ANN-SNN method.}}
\label{fig:ann2snn}
\end{figure}

\definecolor{LightCyan}{rgb}{0.88,1,1}
\definecolor{LightOrange}{rgb}{1,0.85,0.70}
\definecolor{DarkCyan}{rgb}{0,0.8,0.8}
\definecolor{DarkOrange}{rgb}{1,0.50,0.0}
\begin{table}[t]
    \centering
    \caption{\textbf{Ablation study on different diffusion solvers.} $S$ denotes the diffusion timesteps. The best results for each SDM solver are shown in bold.}
    \label{tab:fast_solver}
    \resizebox{1\linewidth}{!}{
    \begin{tabular}{ccccccc}
    \toprule 
      & \multicolumn{6}{c}{CIFAR10 ($32 \times 32$)}  \\ 
     $S$ & 10& 20 & 50 & 100 & 200 & 500  \\ \midrule
     \multicolumn{1}{l}{DDIM}  &48.51& 30.60 & 22.46 & 20.31 & \textbf{18.49} & 19.02\\
     \multicolumn{1}{l}{SDDIM (T=8)} &39.73&21.69 & \textbf{16.02}& 16.27&18.43 &23.93\\
    \multicolumn{1}{l}{SDPM-Solver (T=8)} &30.97&\textbf{30.85} & 31.80& 32.33&32.42 &32.06\\
    \multicolumn{1}{l}{Analytic-SDPM (T=8)} &58.38& 28.31 & 17.35 & 15.38 &13.41 &\textbf{12.95}\\
        \bottomrule
    \end{tabular}
    }
    
\end{table}

\subsection{Ablation Study}
\noindent\textbf{Impact of different components of SDMs.} We first conduct ablation studies on CIFAR-10 to investigate the effects of Temporal-wise Spiking Module (TSM) and Threshold Guidance (TG). As presented in Tab.~\ref{tab:ablation}, we observe that both TSM and TG contribute to improving image quality. 
The optimal FID results are obtained by using TSM and TG simultaneously, achieving an 18.4\% increase compared to Vanilla SDDIM.

\noindent\textbf{Effectiveness of SDM on different solvers.} In Tab.~\ref{tab:fast_solver}, we verify the feasibility and validity of our SDM across various diffusion solvers. SDDIM exhibits a more stable performance depending on sampling steps, while Analytic-SDPM boasts exceptional capabilities, achieving new state-of-the-art performance and surpassing ANN-DDIM results. In conclusion, our SDM proves its proficiency in handling any diffusion solvers, and we believe there remains significant potential for further enhancement of the FID performance utilizing our SDM.

\section{Discussion and Conclusion}
In this work, we propose a new family of SNN-based diffusion models names Spiking Diffusion Models (SDMs) that combine the energy efficiency of
SNNs with superior generative performance. SDMs achieve state-of-the-art results with fewer spike time steps among the SNN baselines and also attain competitive results with lower energy consumption compared to ANNs. SDMs primarily benefit from two aspects: (1) the Temporal-wise Spiking Mechanism (TSM), which enables the synaptic currents of the denoised network SNN-UNet to gather more dynamic information at each time step, as opposed to being governed by fixed synaptic weights as traditional SNNs, and (2) the training-free Threshold Guidance (TG), which can further enhance the sampling quality by adjusting the spike threshold. Nevertheless, one limitation of our work is that the time step of our SNN-UNet is relatively small, leaving the full potential of SDMs unexplored. Additionally, testing on higher-resolution datasets (\eg, ImageNet) should be considered. In future research, we plan to explore further applications of SDMs in the generation domain, \eg, text-image generation, and attempt to combine it with advanced language models to achieve more interesting tasks.

{
\bibliographystyle{IEEEtran}
\bibliography{ref}
}

\begin{IEEEbiography}[{\includegraphics[width=0.8in,height=1.0in,clip,keepaspectratio]{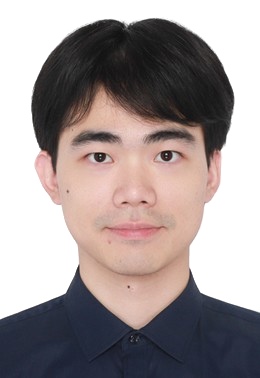}}] {Jiahang Cao} receives the B.S. degree in mathematics from the Sun Yat-sen University. He is currently a Ph.D. student in the Humanoid Computing Lab,  Microelectronics Thrust, The Hong Kong University of Science and Technology (Guangzhou).
His research interests include neuromorphic computing and robot learning.
\end{IEEEbiography}

\begin{IEEEbiography}[{\includegraphics[width=0.8in,height=1.0in,clip,keepaspectratio]{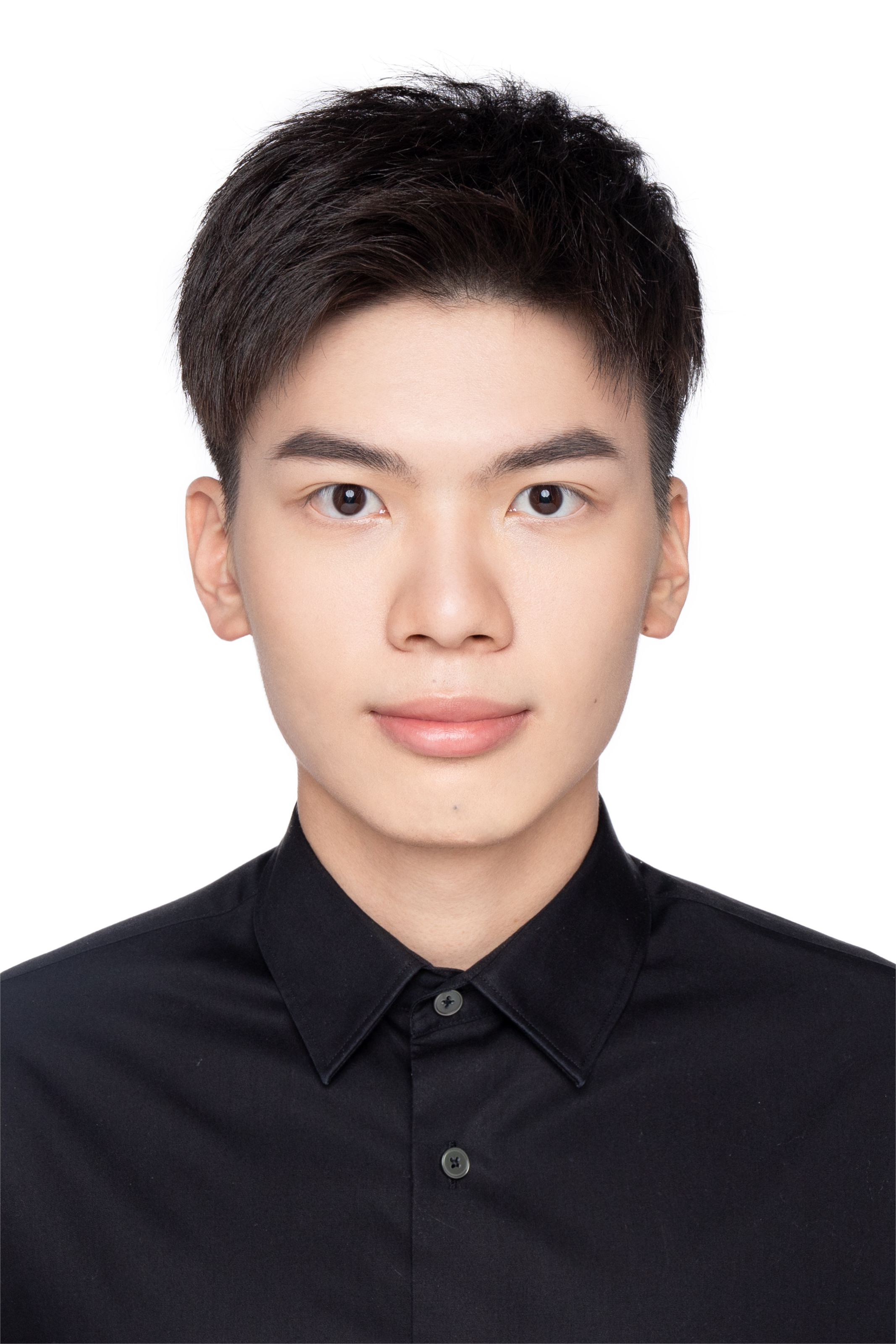}}]{Hanzhong Guo} 
receives the B.S. degree in economics from the Sun Yat-sen University. He is currently a
master student in the Renmin University of China.
His research interests include generative models, computer vision and trustworthy AI.
\end{IEEEbiography}

\begin{IEEEbiography}[{\includegraphics[width=0.8in,height=1.0in,clip,keepaspectratio]{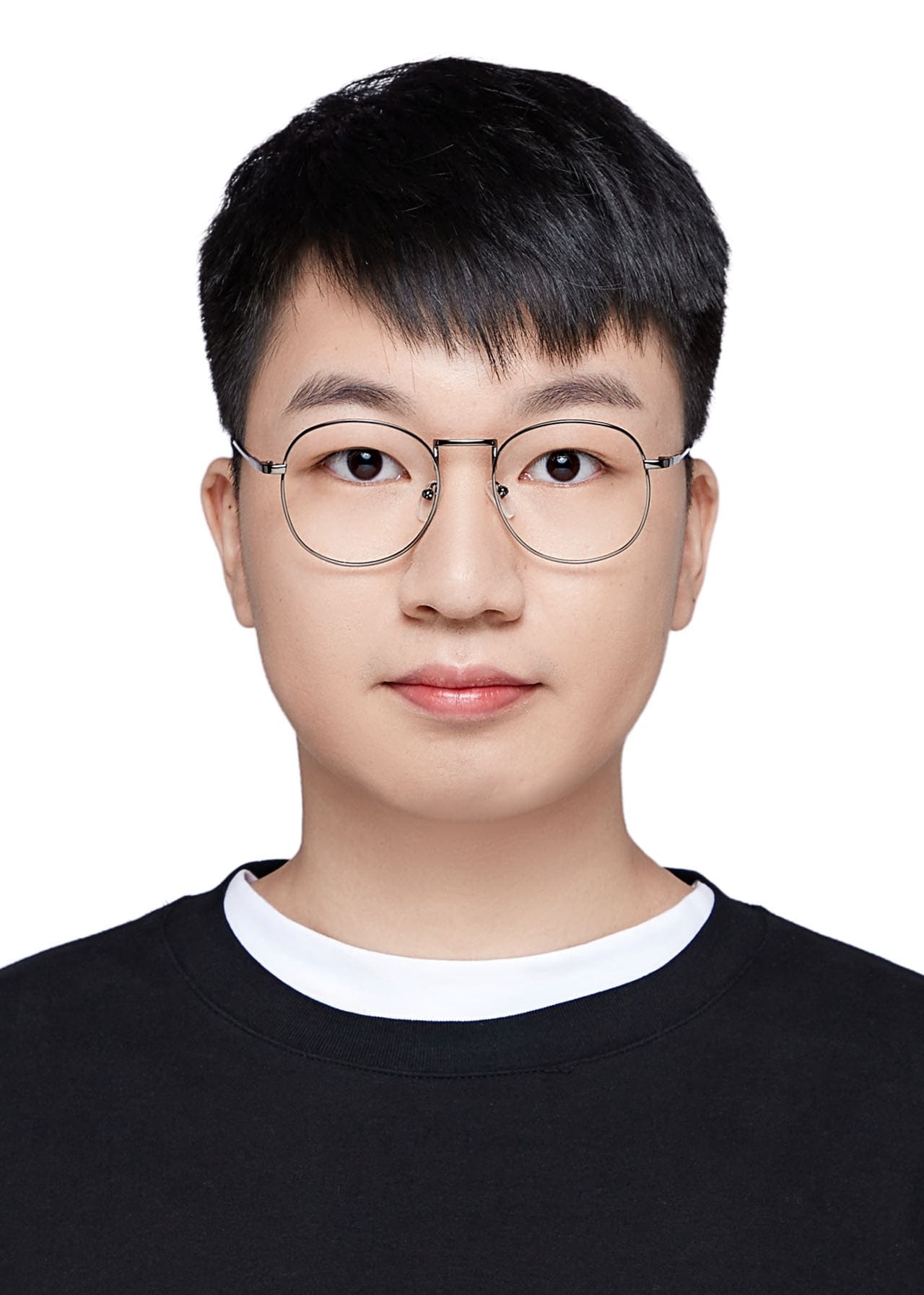}}]{Ziqing Wang} receives the B.S. degree in microelectronics from the Sun Yat-sen University. He is currently a CS Ph.D. student at the North Carolina State University. His research interests include brain-inspired computing, efficient AI, and generative AI.

\end{IEEEbiography}

\begin{IEEEbiography}[{\includegraphics[width=0.8in,height=1.0in,clip,keepaspectratio]{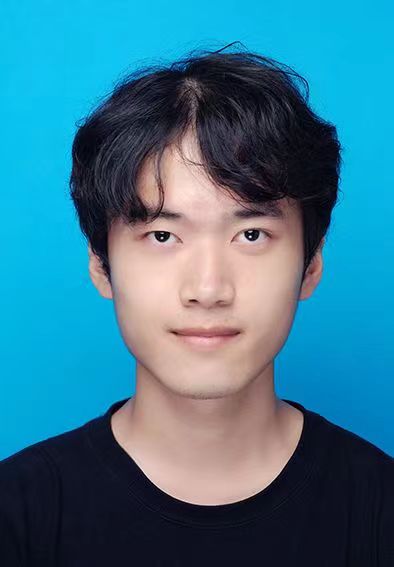}}] {Deming Zhou} is currently an undergraduate in the Department of Information and Computing Sciences, School of Mathematics and Information Science, Guangzhou University.
His research interest is brain-inspired computing.
\end{IEEEbiography}

\begin{IEEEbiography}[{\includegraphics[width=0.8in,height=1.0in,clip,keepaspectratio]{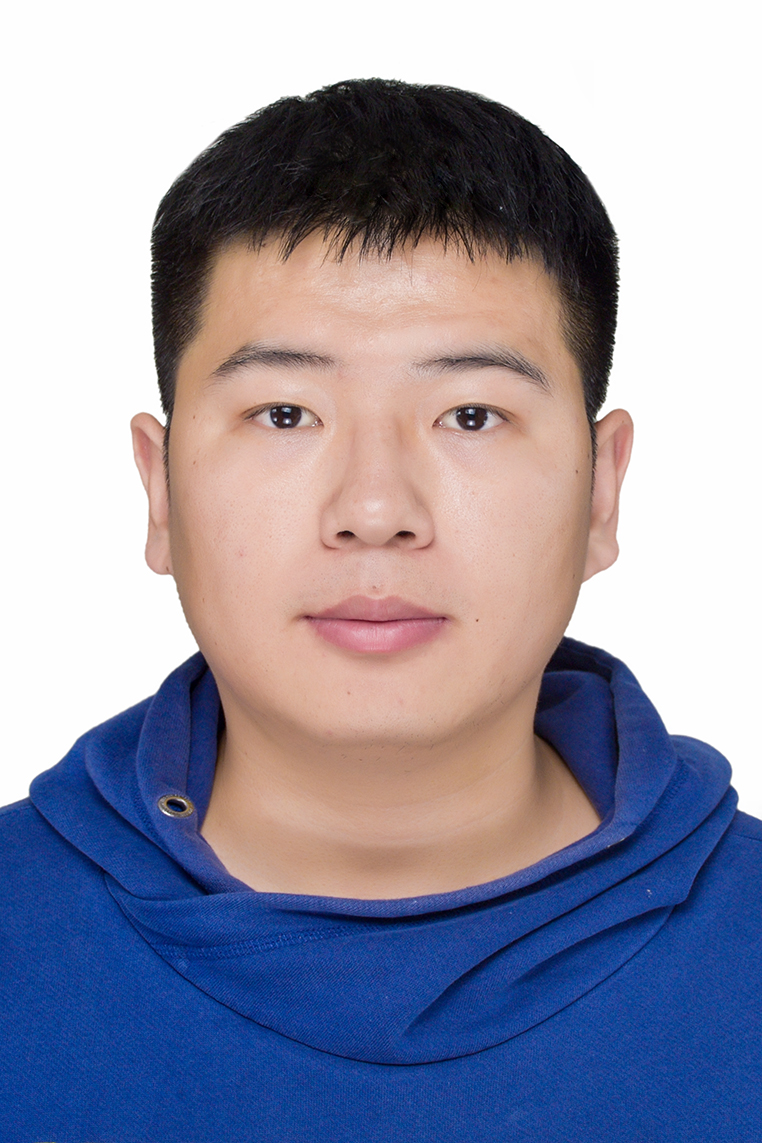}}] {Hao Cheng} is a Ph.D. student in the Humanoid Computing Lab, Microelectronics Thrust, The Hong Kong University of Science and Technology (Guangzhou). His research interests include deep learning robustness, model efficiency,  brain-inspired computing, and machine learning. 

\end{IEEEbiography}

\begin{IEEEbiography}[{\includegraphics[width=0.8in,height=1.0in,clip,keepaspectratio]{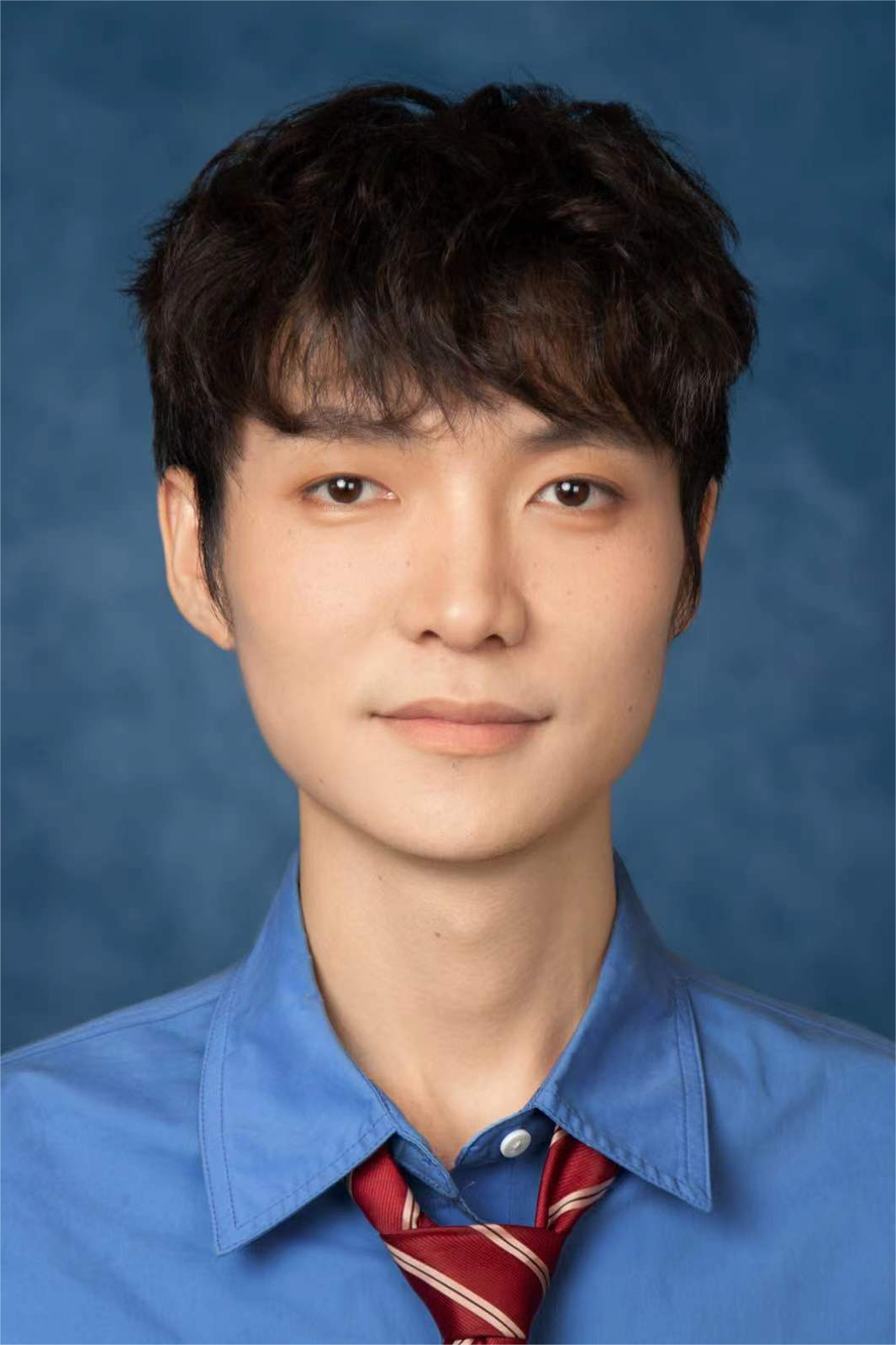}}]{Qiang Zhang} is currently a Ph.D. student in the Humanoid Computing Lab,  Microelectronics Thrust, The Hong Kong University of Science and Technology (Guangzhou).
His research interest is humanoid robots.

\end{IEEEbiography}

\begin{IEEEbiography}[{\includegraphics[width=0.8in,height=1.0in,clip,keepaspectratio]{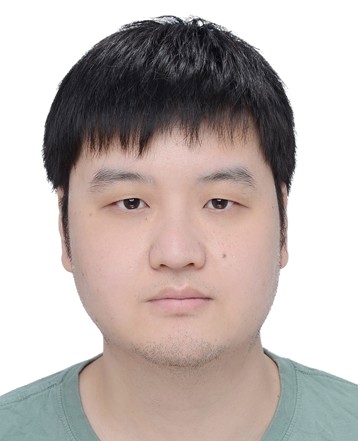}}] {Renjing Xu} is an Assistant Professor at the Microelectronics Thrust, Function Hub of the Hong Kong University of Science and Technology (Guangzhou). He received a B.Eng  (First-class Honors) in photonic system from the Australian National University in 2015 and a Ph.D. in applied physics from Harvard University in 2021 (advised by Prof. Donhee Ham). From 2015 to 2016, he was a visiting scholar in the Photonics Lab at University of Wisconsin-Madison. 
His research interests include human-centered computing and hardware-efficient computing. 
\end{IEEEbiography}

\clearpage
\section*{Appendix}

\subsection{Learning Rules of SNNs}
\label{subsec:appendix_gradient}
In this section, we revisit the overall training algorithm of the deep SNNs. Our goal is to optimize the weights $\mathbf{W}$ in our SNN-UNet as well as the temporal parameter $\mathbf{P}$=$\{p[1],p[2]...,p[T]\}$, where $T$ is the SNN time steps. We follow one of the standard direct-training algorithms Spatio-Temporal Backpropagation (STBP~\cite{wu2018spatio}) to calculate the gradient process.

Consider the noise input $\mathcal{E} = \{\epsilon_1, \epsilon_2, .. \epsilon_N\}$ with batch $N$ and the predicted noise output $\hat{\mathcal{E}} = \{\hat{\epsilon_1}, \hat{\epsilon_2}, .. \hat{\epsilon_N}\}$. The loss function $\mathcal{L}_{mse}$ is defined as:
\begin{align}
    \mathcal{L}_{mse} = \frac{1}{N}\sum^N_{i=1} (\hat{\epsilon_i}-\epsilon_i)^2.
\end{align}
Similar to the previous works~\cite{zheng2021going}, we compute the gradient by unfolding the network
on both spatial and temporal domains. By applying the chain rule, we can get:
\begin{align}
\frac{\partial \mathcal{L}}{\partial W_{ij}^{l}} &=  \frac{\partial \mathcal{L}}{\partial u_i^{l}[t]}\frac{\partial u_i^{l}[t]}{\partial W_{ij}^{l}} = \frac{\partial\mathcal{L}}{\partial u_i^{l}[t]}o_j^{l-1}[t] ,\\
\frac{\partial \mathcal{L}}{\partial p^{l}[t]} &= \sum_i \frac{\partial \mathcal{L}}{\partial u_i^{l}[t]}\frac{\partial u_i^{l}[t]}{\partial p^{l}[t]},
  \end{align}
where $W_{ij}^{l}$ denotes the synaptic weight of SNNs between the $i$-th neuron and $j$-th neuron. $u_i^{l}[t]$ and $o_i^{l}[t]$ denotes the membrane potential and spikes of the $i$-th neuron at layer $l$ of time $t$, respectively. By applying the chain rule, $\frac{\partial \mathcal{L}}{\partial u_i^{l}[t]}$ and $\frac{\partial \mathcal{L}}{\partial o_i^{l}[t]}$ can be computed by:

\begin{align}
\frac{\partial \mathcal{L}}{\partial u_i^{l}[t]} &= \frac{\partial \mathcal{L}}{\partial o_i^{l}[t]}\frac{\partial o_i^{l}[t]}{\partial u_i^{l}[t]}+\frac{\partial \mathcal{L}}{\partial u_i^{l}[t+1]}\frac{\partial u_i^{l}[t+1]}{\partial u_i^{l}[t]},\\
\frac{\partial \mathcal{L}}{\partial o_i^{l}[t]} &= \sum_{j}\frac{\partial\mathcal{L}}{\partial u_j^{l+1}[t]}\frac{\partial u_j^{l+1}[t]}{\partial o_i^{l}[t]}+\frac{\partial \mathcal{L}}{\partial u_i^{l}[t+1]}\frac{\partial u_i^{l}[t+1]}{\partial o_i^{l}[t]}.
\end{align}

However, due to the non-differentiable spiking activities, $\frac{\partial o[t]}{\partial u[t]}$ can not be computed directly. To address this problem, the surrogate gradient method is used to smooth out the step function by the following equation:
\begin{align}
    \frac{\partial o[t]}{\partial u[t]} &=  \frac{1}{a}sign(|u[t]-\vartheta_{\textrm{th}}|<\frac{a}{2}).
\end{align}

\subsection{Different Solvers for Diffusion Models}
The detail of integrating the proposed model with various inference solvers is that: 
The sampling process can be divided into multiple discretization steps and our proposed spiking UNet is trained to predict noise in each discretization step. 
Different solvers essentially optimize the discretization method, where each step of discretization requires inference based on the pre-trained noise prediction model.
Therefore, for different diffusion solvers, the pre-trained noise prediction network is fixed. We introduce more explanations as follows. 

Diffusion models incrementally perturb data through a forward diffusion process and subsequently learn to reverse this process to restore the original data distribution. Formally, let $x_0 \in \mathbb{R}^n$ represent a random variable with an unknown data distribution $q(x_0)$. The forward process, denoted as $x_t$, where $t\in [0,1]$ and indexed by time $t$, perturbs the data by adding Gaussian noise to $x_0$
\begin{eqnarray}
\label{eq:dpmf}
\label{eq:forward}
     q(x_t|x_0) = \mathcal{N}(x_t|a(t)x_0,\sigma^2(t)I).
\end{eqnarray}
In general, the function $a(t)$ and $\sigma(t)$ are selected so that the logarithmic signal-to-noise ratio {\small $\log \frac{a^2(t)}{\sigma^2(t)}$} decreases monotonically with time $t$, causing the data to diffuse towards random Gaussian noise~\cite{kingma2021variational}. 
Furthermore, it has been demonstrated by \cite{kingma2021variational} that the following SDE shares an identical transition distribution $q_{t|0}(x_t|x_0)$ with Eq.~\eqref{eq:dpmf}:
\begin{eqnarray}
     d x_t = f(t)x_td t + g(t)d \omega  ,\quad x_0\sim q(x_0),
\end{eqnarray}
where $\omega \in \mathbb{R}^n$ is a standard Wiener process and 
\begin{eqnarray}
    f(t)=\frac{d \log a(t)}{d t} ,\quad g^2(t)=\frac{d \sigma^2(t)}{d t}-2\sigma^2(t)\frac{d \log a(t)}{d t} .
\end{eqnarray}
Let $q(x_t)$ be the marginal distribution of the above SDE at time $t$. Its reversal process can be described by a corresponding continuous SDE which recovers the data distribution:
\begin{eqnarray}
    d x = \left [ f(t)x_t-g^2(t) \nabla_{x_t} \log q(x_t) \right ] d t + g(t)d \bar{\omega} ,
\end{eqnarray}
where $x_1\sim q(x_1)$, $\bar{\omega} \in \mathbb{R}^n$ is a reverse-time standard Wiener process. Therefore, to sample the data, it is need to replace the $ \log q(x_t) $ with a learned score network $s_\theta(x_t,t)$ or noise network ($-\frac{1}{\sigma^2(t)}\epsilon_\theta(x_t,t)$) and discretize the reverse SDE or corresponding ODE in Eq.~\eqref{pfode} to obtain the generated samples $\hat{x}_0$.
\begin{equation}
\label{pfode}
    \frac{dx_t}{dt} = f(t)x_t - \frac{1}{2}g^2(t)\nabla_{x}\log q_t(x_t), \quad x_1\sim q(x_1).
\end{equation}
For the SDE-based solvers, their primary objective lie in decreasing discretization error and therefore minimizing function evaluations required for convergence during the process of discretizing Eq.~\eqref{eq:continuous_back}.
Discretizing the reverse SDE in Eq.~\eqref{eq:continuous_back} is equivalent to sample from a Markov chain $p(x_{0:1})=p(x_1)\prod_{t_{i-1},t_i \in S_t}p(x_{t_{i-1}}|x_{t_{i}})$ with its trajectory $S_t=[0,t_1,t_2,...,t_i,..,1]$.
Song \etal~\cite{song2020score} proves that the conventional ancestral sampling technique used in the DPMs~\cite{ho2020denoising} that models $p(x_{t_{i-1}}|x_{t_{i}})$ as a Gaussian distribution, can be perceived as a first-order solver for the reverse SDE in Eq.~$\eqref{eq:continuous_back}$. 
Bao \etal~\cite{bao2022analytic} finds that the optimal variance of $p(x_{t_{i-1}}|x_{t_{i}})\sim \mathcal{N}(x_{t_{i-1}}|\mu_{t_{i-1}|t_i},\Sigma_{t_{i-1}|t_i}(x_{t_i}))$ is 
\begin{align}
\label{eq:optimalvariance}
& \Sigma^*_{t_{i-1}|t_i}(x_{t_i})=  \lambda_{t_i}^2 +  \nonumber \\ 
& \gamma_{t_i}^2 \frac{\sigma^2(t_i)}{\alpha(t_i)} 
\left(1-\mathbb{E}_{q(x_{t_i})}\left[\frac{1}{d} \left \| \mathbb{E}_{q(x_{0}|x_{t_i})}[\epsilon_\theta(x_{t_i},t_i)] \right \|_2^2\right]\right),
\end{align}
where, $\gamma_{t_i} = \sqrt{\alpha(t_{i-1})} - \sqrt{\sigma^2(t_{i-1}) - \lambda^2_{t_i}} \sqrt{\frac{\alpha(t_i)}{\sigma^2(t_i)}}$, $\lambda_{t_i}^2$ is the variance of $q(x_{t_{i-1}}|x_{t_i},x_0)$. 
AnalyticDPM~\cite{bao2022analytic} offers a significant reduction in discretization error during sampling and achieves faster convergence with fewer steps. 
Besides the learned noise network $\epsilon_\theta(x_t,t)$, to use the the AnalyticDPM to sampling, it is required to obtain the statistic value $h(t_i)=\mathbb{E}_{q(x_{t_i})}\left[\frac{1}{d} \left \| \mathbb{E}_{q(x_{0}|x_{t_i})}[\epsilon_\theta(x_{t_i},t_i)] \right \|_2^2\right]$. Therefore, we can pre-calculate the $h(t_i)$ for each dataset such as CIFAR-10 via using the SOTA pre-trained diffusion models and use the  $h(t_i)$ to calculate the optimal variance of reverse transition kernel in sampling, which won't affect the speed of sampling while significant improving the generation performance of spiking diffusion models.

In another setting, as for the ODE-based solvers, their primary objective lie in decreasing discretization error by using the higher-order solvers for the reverse ODE.
Some researches figure out that the first of the two terms of the reverse ODE can be computed explicitly, while only the term with the noise network needs to be estimated, hence the following equation, 
\begin{equation}
\label{eqn:variation_of_constants}
    x_t = e^{\int_s^t f(\tau)d\tau}x_s + \int_s^t \left(e^{\int_\tau^t f(r)d r}\frac{g^2(\tau)}{2\sigma_\tau} \epsilon_\theta(x_\tau,\tau)\right)d\tau.
\end{equation}
In order to decrease the discretization error, it can use higher-order estimates to approximate the terms followed by the noise network. With the first-order estimation, we obtain the following equation which is the same with the DDIM~\cite{song2020denoising},
\begin{equation}
\label{eqn:1st}
    \tilde{x}_{t_i} = \frac{\alpha_{t_i}}{\alpha_{t_{i-1}}} \tilde{x}_{t_{i-1}} - \sigma_{t_i} (e^{h_i} - 1)\epsilon_\theta(\tilde{x}_{t_{i-1}},t_{i-1}),
\end{equation}
where $h_i=\lambda_{t_i} - \lambda_{t_{i-1}}$. While approximating the first k terms of the Taylor expansion needs additional intermediate points between t and s, more details in DPM-Solver~\cite{lu2022dpm}.

However, since the estimation error in spiking diffusion models can't be ignored, the performance of ODE-based solvers is worse than the performance of SDE-based solvers, which is shown in the experiments section. Meanwhile, the parameters in different solvers are set as default setting.

\subsection{Hyperparameter Settings}
As illustrated in Tab.~\ref{tab:hyper}, we have presented all the hyperparameters of the Spiking Diffusion Model (SDM) to ensure the reproducibility of our results. We are committed to transparency in our research and will open source our code in the future.

\begin{table}[ht]
    \caption{Hyperparameters of the spiking diffusion models.}
    \label{tab:hyper}
    \centering
    \resizebox{\linewidth}{!}{
    \begin{tabular}{lc}
        \toprule
        \textbf{{Hyperparameter}}          & \textbf{{Value}}         \\
        \midrule
       \textit{{Spiking UNet hyperparameter}}       &      \\
        \midrule
        {Diffusion timesteps}     & {1000}          \\
        {Base channel dimension}  & {128}           \\
        {Time embedding dimension} & {512}          \\
        {Channel multipliers}     & {[1, 2, 2, 4]}     \\
        {Dropout rate}               & {0.1}           \\
        {Spiking timesteps}       & {4 or 8}        \\
        \midrule
        \textit{{Diffusion sampler hyperparameter}}      &      \\
        \midrule
        {Beta\_1}                 & {1e-4}          \\
        {Beta\_T}                 & {0.02}          \\
        {Sampler type}            & {DDPM, DDIM, DPM-Solver, Analytic-DPM} \\
        \midrule
        \textit{{Training hyperparameter}}       &      \\
        \midrule
        {Batch size}              & {128*GPUs}            \\    
        {Training steps}          & {500,000}              \\
        {Gradient clip}           & {1.0}                  \\
        \bottomrule
    \end{tabular}}
\end{table}

{
\subsection{Decided Threshold Voltage}
{In this section, we provide more details of deciding the threshold voltage in the spiking diffusion models.}
As shown in the paper, the change of score incurred by the change of the threshold voltage can be divided into two terms, one for the original estimated score, another is the rectified term $c_{\theta}(x_t,t)$. 
\begin{align}
&s_{\theta}(x_t,t,\vartheta_{\textrm{th}}') \nonumber \\
&\approx s_{\theta}(x_t,t,\vartheta_{\textrm{th}}^0)+\frac{\mathrm{d} s_{\theta}(x_t,t,\vartheta_{\textrm{th}})}{\mathrm{d} \vartheta_{\textrm{th}}} \mathrm{d} \vartheta_{\textrm{th}} + O(\mathrm{d} \vartheta_{\textrm{th}}) \nonumber \\
&\approx s_{\theta}(x_t,t)+ s'_{\theta}|_{\vartheta_{\textrm{th}}^0} \mathrm{d} \vartheta_{\textrm{th}} + O(\mathrm{d} \vartheta_{\textrm{th}})
 \nonumber \\
&\approx s_{\theta}(x_t,t)+c_{\theta}(x_t,t),
\end{align}
Therefore, due to the estimation error of the noise network, a discrepancy exists between the estimated score $s_\theta(x_t,t)$ and the true score $s(x_t,t)$. Since the true score is unknown, it is impossible to calculate the actual discrepancy and thus $c_\theta(x_t,t)$.
Furthermore, even if the gap between the true score and the estimated score is calculated, solving the corresponding differential equation is challenging, and it is uncertain whether a solution exists
However, our findings are consistent with those of Ho et al.~\cite{ho2022classifier}, who demonstrated that generation quality can be improved by introducing classifier-free guidance. This improvement is essentially due to the gap between the estimated score and the true score, where an artificially introduced correction term enhances the results. 
In the future, we will explore methods to find the optimal threshold voltage.
}



{
\subsection{Effectiveness of the Temporal-wise Spiking Mechanism}}
{To better visualize the performance improvement brought by the TSM module, we have provided the generation results of CIFAR-10 using SDDIM with and without the TSM module. Here we used DDIM instead of DDPM for this comparison since DDIM operates based on Ordinary Differential Equations (ODEs), which ensure deterministic and consistent generation results. In contrast, DDPM relies on Stochastic Differential Equations (SDEs), which introduce randomness in the generation process, leading to variability in the output images and making direct comparisons challenging.}

\begin{figure}[h]
	\setlength{\tabcolsep}{1.0pt}
	\centering
	\begin{tabular}{c}
        \includegraphics[width=.49\textwidth]{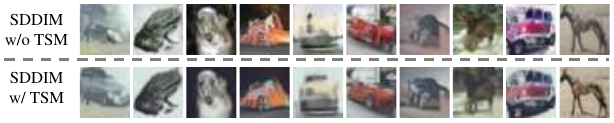} 
	\end{tabular}
	\caption{{\textbf{Comparisons of the generation results with/without using the TSM method in CIFAR-10.}  }}
	\label{fig:vis_appendix}
\end{figure}

{The results in Fig.~\ref{fig:vis_appendix} demonstrate a significant improvement in the quality of the generated images with the TSM module. The contours of the images are more pronounced, the backgrounds are clearer, and the texture details are richer compared to those without the TSM module, thereby proving the effectiveness of TSM.} 

\end{document}

%% file: Alg_tsm.tex
\renewcommand{\algorithmicrequire}{\textbf{Input:}}
\renewcommand{\algorithmicensure}{\textbf{Output:}}

\begin{algorithm}[t]\small
\caption{The Learning Pipeline of the SDM.}
\label{alg:pipeline}
    \begin{algorithmic}[1]
        \REQUIRE the number of finetune iterations $I_{ft}$, the number of SNN layers $L_{snn}$, spiking time steps $T_{snn}$, encoding layer $\mathscr{E}$, diffusion process $\mathscr{D}$, decay factor $\gamma$, threshold $\vartheta_{\textrm{th}}$, learning rate $\alpha$
        \ENSURE Optimized $W$, $p$
        \vspace{0.05in}
        \hrule
        \vspace{0.05in}
        \STATE Training spiking diffusion models with the standard pre-spike residual block (Sec.~\ref{subsec:pre_spike}). \hfill $\triangleright$ \textbf{Stage 1}
        \vspace{0.05in}
        \STATE Prepare to finetune the spiking diffusion models with the TSM residual block (Sec.~\ref{subsec:tsm}).  \hfill $\triangleright$ \textbf{Stage 2}
        \vspace{0.05in}
        \STATE Inherit pretrained weight $W$ and initialize temporal parameter $p$ with 1.0.
        \vspace{0.05in}
        \FOR{$i=1$ to $I_{ft}$}
        \vspace{0.05in}
        \STATE Initialize Gaussian noise $\epsilon \sim \mathcal{N}(0,\mathcal{\textbf{I}})$. 
        \vspace{0.05in}
        \STATE Calculate the input spikes $S^0$ = $\mathscr{E}(\epsilon)$.
        \vspace{0.05in}
        \FOR{$l=1$ to $L_{snn}$}
        \vspace{0.05in}
            \FOR{every neuron $i$ in layer $l$}
            \vspace{0.05in}
                \FOR{$t=1$ to ${T_{snn}}$}
                \vspace{0.05in}
                \STATE $U^l_i[t]$ = $e^{\frac{1}{\gamma}}\cdot{U}_i^l[t-1](1-S_i^l[t-1]) + \widehat{I}_i^{l-1}[t]$
                \vspace{0.05in}
                \STATE $S^l_i[t]$ = $\Theta ({U}_i^l[t]-\vartheta_{\textrm{th}})$
                \vspace{0.05in}
                \STATE $I^{l}_i[t]$ = $\mathbf{W}_i^{l+1}S^{l}_i[t]$
                \vspace{0.05in}
                \STATE $\widehat{I}^{l}_i[t]$ = $I^{l}_i[t] \times p[t]$ 
                \vspace{0.05in}
                \ENDFOR
                \vspace{0.05in}
            \ENDFOR
            \vspace{0.05in}
        \ENDFOR
        \vspace{0.05in}
        \STATE $\epsilon_{out}$ = $\mathscr{D}({U}^L[1:T_{snn}])$
        \vspace{0.05in}
        \STATE $\{W, p\} \leftarrow \{W, p\} - \alpha \nabla_{\{W, p\}} \cdot \mathcal{L}_{MSE}(\epsilon,\epsilon_{out})$
        \vspace{0.05in}
        \ENDFOR
    \end{algorithmic}
\end{algorithm}

\setlength{\textfloatsep}{5pt}